\documentclass[10pt,twocolumn,letterpaper]{article}

\usepackage[final]{cvpr}      

\usepackage{graphicx}
\usepackage{amsmath}
\usepackage{amssymb}
\usepackage{booktabs}
\usepackage{multirow}
\usepackage{makecell}

\usepackage[pagebackref,breaklinks,colorlinks]{hyperref}

\usepackage[capitalize]{cleveref}
\crefname{section}{Sec.}{Secs.}
\Crefname{section}{Section}{Sections}
\Crefname{table}{Table}{Tables}
\crefname{table}{Tab.}{Tabs.}

\newtheorem{lemma}{Lemma}

\def\cvprPaperID{253} 
\def\confName{CVPR}
\def\confYear{2022}

\begin{document}
	
	\title{IMFNet: Interpretable Multimodal Fusion for Point Cloud Registration}
	
	\author{Xiaoshui Huang$^{[1]}$, Wentao Qu$^{[2]}$, Yifan Zuo$^{[2]\thanks{Corresponding author}}$, Yuming Fang$^{[2]}$, Xiaowei Zhao$^{[3]}$ \\
		$[1]$ Image X Institute, University of Sydney,
		$[2]$ Jiangxi University of Finance and Economics,
		$[3]$ Sany
}
\maketitle
\begin{abstract}
	The existing state-of-the-art point descriptor relies on structure information only, which omit the texture information. However, texture information is crucial for our humans to distinguish a scene part. Moreover, the current learning-based point descriptors are all black boxes which are unclear how the original points contribute to the final descriptor. In this paper, we propose a new multimodal fusion method to generate a point cloud registration descriptor by considering both structure and texture information. Specifically, a novel attention-fusion module is designed to extract the weighted texture information for the descriptor extraction. In addition, we propose an interpretable module to explain the original points in contributing to the final descriptor. We use the descriptor element as the loss to backpropagate to the target layer and consider the gradient as the significance of this point to the final descriptor. This paper moves one step further to explainable deep learning in the registration task. Comprehensive experiments on 3DMatch, 3DLoMatch and KITTI demonstrate that the multimodal fusion descriptor achieves state-of-the-art accuracy and improve the descriptor's distinctiveness. We also demonstrate that our interpretable module in explaining the registration descriptor extraction.
\end{abstract}

\section{Introduction}
\label{sec:intro}
Point cloud registration is a technique that aims to estimate the transformation matrix (rotation and translation) between two point clouds. This technique played a critical role in many applications, including robotics and augmented reality\cite{huang2021comprehensive}. Among the existing registration methods\cite{choy2019fully,huang2021predator,ao2021spinnet,huang2020feature,horache20213d}, descriptor-based methods \cite{ao2021spinnet,horache20213d} are an important category and achieve the state-of-the-art accuracy in the large-scale real-world datasets (e.g., 3DMatch \cite{zeng20173dmatch}). The distinctiveness of the 3D point descriptor dominates the performance of these descriptor-based registration methods. 

\begin{figure}[t]
	\includegraphics[width=\linewidth,height=6cm]{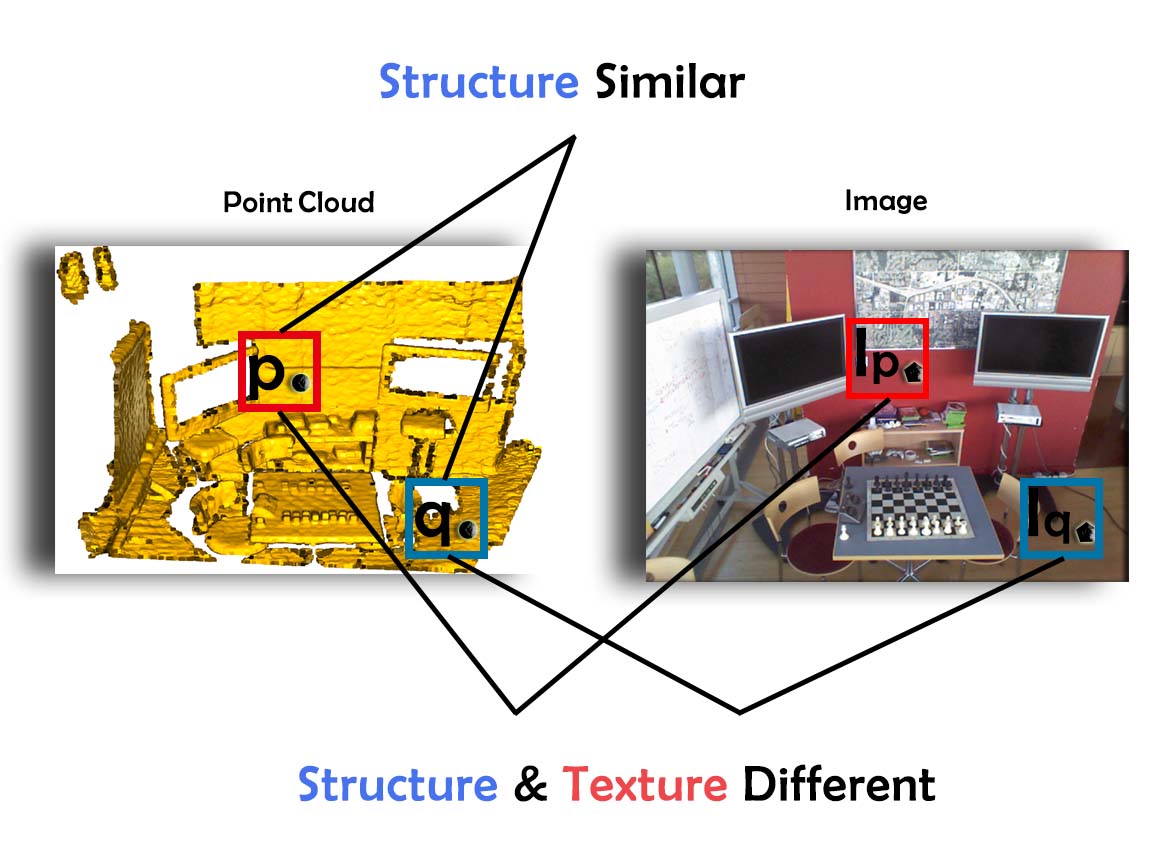}
	\caption{A visual example to show that descriptors of non-matched points \textbf{p} and \textbf{q} are similar if considering the structure only (top). However, descriptors are discriminative when considering structure and texture information (bottom).}
	\label{f1}
\end{figure}

Most of the current 3D descriptors utilize the structure information to describe the points \cite{rusu2009fast,choy2019fully,huang2021predator,ao2021spinnet}. 
However, repeatable and ambiguous structures widely exist in point clouds, such as floor, wall, and ceiling are all planes (see Figure \ref{f1} as an example). These repeatable and ambiguous structure information will largely impact the descriptor distinctiveness. Consequently, the correspondences estimated by comparing the structure-only point descriptors contain significant outliers. The existing published literature \cite{choy2019fully,huang2021predator,ao2021spinnet}  has demonstrated this phenomenon that the feature match recall drops a lot when the inlier threshold increases to 0.2. Moreover, the current descriptors' neural networks are all black boxes. We never know how the structure information contributes to the final descriptor. Without knowing the internal mechanism of the descriptor extraction process, it is difficult to understand the reason for its failures or success from new testing datasets. This paper aims to improve the distinctiveness and unfold the black box for 3D point descriptor learning.

To improve the distinctiveness of point descriptors, we propose a new multimodal fusion method to learn the 3D point descriptor by fusing the structure information of the point cloud and the texture information of its corresponding image.
Our motivation lies in our humans usually considering both texture and structure when we watch a scene and discriminate two parts—for example, red wall($I_p$) and yellow floor($I_q$) (see Figure \ref{f1} as an example). Moreover, the current vision system in the intelligent agents (e.g., autonomous cars and home robotics) usually contains both point cloud sensors and image sensors. Data acquisition of both point clouds and images becomes widely affordable.

Specifically, our multimodal fusion method is an encoder and decoder architecture based on FCGF \cite{choy2019fully}. Inspired by the transformer\cite{vaswani2017attention}, a new cross attention module is developed to extract the weighted texture information for each point after the encoder module. Then, we concatenate the texture and structure information and feed them into the decoder module for the final descriptor learning. 

Moreover, we move one step further to unfold the black box for the 3D descriptor learning. We designed an interpretable module, descriptor activation mapping (DAM), which interprets how the neighbour points are involved in the descriptor extraction. With the interpretable module, the descriptor internal generation process is presented to us. Our interpretable module is inspired by Grad-CAM\cite{selvaraju2017grad} but different to Grad-CAM with several critical improvements and theoretical analysis specifically for the registration task. The reasons for these improvements lie in two aspects: (1) the Grad-CAM can be applied to ordinary 3D CNN but fails to sparse tensor convolution. (2) the Grad-CAM requires a class label (e.g., dog or cat) to calculate the category-specified gradient, but the registration task does not contain such class labels.

Specifically, our DAM introduces a novel method to calculate a heat map that consists of the significance of the input points contributing to the output in the target layer. We use the descriptor's channel value as the loss to backpropagate the gradient into the target layer, which does not require class labels. To interpret the multimodal fusion descriptor, we considers the last layer as the target layer and constructs a heat map based on the gradient addition of all the descriptor's channels.

The main contributions of this paper could be summarized as
\begin{itemize}
	\item A novel multimodal fusion method is proposed to learn 3D point descriptors with texture and structure information. Our method will improve the descriptor's distinctiveness.
	
	\item An interpretable module is proposed to unfold the black box of the 3D point descriptor neural network. This module will interpret how the neighbour points in contributing the final descriptor.
	
	\item Comprehensive experiments demonstrate that  the proposed multimodal fusion descriptor achieves the state-of-the-art performance on both indoor and outdoor datasets.
	
\end{itemize}

\section{Related works}
Our work builds on prior work in several domains: 3D descriptor and visual explanations.

\subsection{3D descriptor}
Before deep learning was prevalent in 3D computer vision, many handcrafted descriptors were proposed to utilize the structure information (e.g., edge, face, normal) to describe the points, such as FPFH \cite{rusu2009fast} and ESF  \cite{wohlkinger2011ensemble}. Several pieces of literature consider the texture and structure information into separate descriptors and combine them into an optimization process to solve the registration \cite{lin2016color, park2017colored}. For example, ColorICP \cite{park2017colored} improved the ICP by adding a 3D color objective. Recently, \cite{yang2020color} designed a hybrid descriptor by concatenating the spatial coordinates and colour moment vector.

After deep learning is introduced into the point cloud registration task, many learning descriptors \cite{zeng20173dmatch,choy2019fully,bai2020d3feat}  designed neural networks to learn the descriptor by utilizing the neighbour structure information of one point cloud. The recent PREDATOR \cite{huang2021predator} designs a transformer module for learning the point descriptor by considering the neighbour structure information of paired point clouds. Because of information fusion of paired point clouds, PREDATOR improves the descriptor's discriminative. SpinNet \cite{ao2021spinnet} projected the point clouds into a spherical space and used spherical convolution operations to extract the structural information as the point descriptor. MS-SVConv \cite{horache20213d} designed a multi-scale framework to learn the descriptor by exploring the multi-scale structure information in describing the points.

However, these descriptors are still facing a challenge in distinctively representing the 3D points. Although handcraft features utilize colour information, the strategies are straightforward: they use it directly or concatenate it with spatial coordinates. The recent state-of-the-art descriptor has not considered the image texture information. However, the texture information is crucial in improving the descriptor distinctiveness. This paper aims to improve the distinctiveness of descriptors by integrating both texture and structure information.

\begin{figure*}[htp]
	\centering
	\includegraphics[width=\textwidth, height=5.5cm]{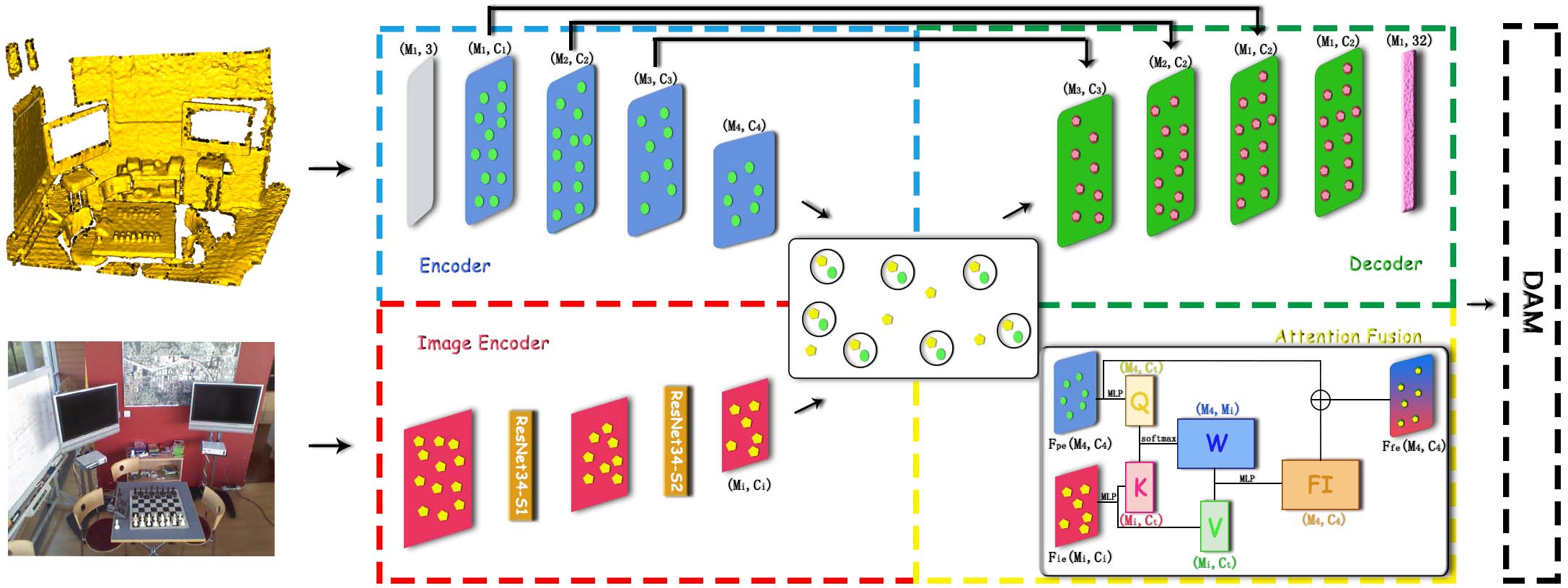}
	\caption{The network architecture of the proposed IMFNet. The input is a point cloud and an image, and the output is a point descriptor. Inside the attention-fusion module, $W$ is the weight matrix, $FI$ is the point texture feature. Then, the point structure feature ($F_{pe}$) and point texture feature ($FI$) are concatenated as an input to the decoder module to get the output descriptor. Descriptor activation mapping (DAM) interprets how the neighbour points in contributing the final descriptor.}
	\label{fig:2}
\end{figure*}

\subsection{Visual explanations}
Interpretable learning has endured several developments in 2D image fields. There are mainly two categories in the area of convolution neural networks. 1)  those interpret how the intermediate layers represent in the real world, and 2) those try to map back the output in the input space to visualize which parts contribute to the output.

One example of the first category is the deep generator network (DQN) \cite{nguyen2016synthesizing}. DQN generates synthetic images for each neuron and reveals the features learned by each neuron in an interpretable way.  \cite{dabkowski2017real} proposed a method to build a saliency map related to the output and explain the relationship between inputs and outputs that the model learned. 
The complete review is beyond the scope of this paper. Please refer the survey \cite{arrieta2020explainable} for more information.

The proposed interpretable module belong to the second category. One wide-known example of the second category is Grad-CAM \cite{selvaraju2017grad}, which flows the gradient back into the convolutional layers to produce a localization map highlighting the important regions in the input for predicting the output. \cite{yang2018visual} extends Grad-CAM to 3D-CNN in solving the Alzheimer’s disease classification. \cite{ghadai2018learning} extends the Grad-CAM to recognize difficult-to-manufacture drilled holes in a complex CAD geometry. 

However, the existing interpretable methods are all focused on classification tasks. None of them focuses on registration tasks that are also an important branch of computer vision. Moreover, the Grad-CAM works on ordinary CNN but faces difficulty in directly applying to sparse tensor convolution. The reason is that the sparse tensor framework has no direct feature map gradient, which is critical for Grad-CAM \cite{choy2019fully}. Nevertheless, sparse tensor convolution has been widely used in the point cloud registration problem. In this paper, we aim to build an interpretable method for the registration problem.

\section{Algorithm: IMFNet}
The overall architecture of the proposed \textbf{i}nterpretable \textbf{m}ultimodal \textbf{f}usion \textbf{net}work (IMFNet) is surprisingly simple and depicted in Figure \ref{fig:2}. It follows the standard UNet architecture with four main components: the encoder module including point encoder and image encoder, attention-fusion module, decoder module and interpretable descriptor activation mapping (DAM) module. 

IMFNet is implemented in Minkowski Engine \cite{choy20194d} and PyTorch that provides the sparse tensor convolution and common CNN backbone architecture implementation with just a few hundred lines. The new attention-fusion module can be implemented within 50 lines.
We hope that the simplicity of our method will attract more researchers to utilize the multimodal information and develop interpretable learning algorithms on the registration problem.

\subsection{Encoder}
The point Encoder follows the FCGF \cite{choy2019fully} to use four sparse tensor convolution layers. The input is $P \in \mathbb{R}^{M_1\times3}$ and the output is $F_{pe} \in \mathbb{R}^{M_4\times C_4}$. The image decoder is a pre-trained ResNet34. The input is $I \in \mathbb{R}^{W\times H \times 3}$ and the output is the feature of the second stage $F_{ie} \in \mathbb{R}^{M_i\times C_i}$. In our algorithm, we consider the whole pixels as one dimension $M_i = H/8*W/8$, which is similar to point dimension.

\subsection{Attention-Fusion}
The goal of attention-fusion module is to extract the texture information for each point to increase the descriptors' distinctiveness. Inspired by the Transformer \cite{vaswani2017attention}, our attention-fusion module follows the cross-attention style. The input is $F_{pe} \in \mathbb{R}^{M_4\times C_4}$ with rich structural information and $F_{ie} \in \mathbb{R}^{M_i\times C_i}$ with texture information. $M_4$ is the number of abstract points of the point encoder, and $M_i$ is the number of abstract pixels of the image encoder. The output of attention-fusion is  $F_{fe} \in \mathbb{R}^{M_4\times C_4}$, which fuses weighted texture information and structure information for each abstract point.

Specifically, the $F_{pe}$ and $F_{ie}$ firstly go through a one-layer MLP. Then, the output of $F_{pe}$ is considered as the query array $Q\in R^{M_4 \times C_t}$, the output of $F_{ie}$ is regarded as key array $K \in R^{M_i \times C_t}$ and value array $V \in R^{M_i \times C_t}$. $C_t$ is the MLP output dimension. As shown in Figure \ref{fig:2}, the $W\in R^{M_4\times M_i}=softmax(\frac{QK^T}{\sqrt{C_t}})$ is the weight matrix that represents the weight of each pixel' texture information that could contribute to describing each point. Then, the $FI\in \mathbb{R}^{M_4\times C_4}$ is the point texture feature that is calculated with one-layer MLP. Mathematically, the point texture feature $FI$ could be calculated,  
\begin{eqnarray}
	FI =MLP(W*V)
\end{eqnarray}

Finally, we conduct an element-wise addition between the point texture feature $FI$ and the spatial structure feature $F_{pe}$ to fuse multimodal information. Mathematically, the fused encoder feature ($F_{fe}$) is calculated as
\begin{eqnarray}
	F_{fe}^{ij} =F_{pe}^{ij}+FI^{ij}, \forall i\in[1,M_4],\forall j\in[1,C_4] 
\end{eqnarray}

\subsection{Decoder} 
Following FCGF \cite{choy2019fully}, we use four sparse tensor transpose convolution to decode the point descriptors. The key difference is that the input of the decoder module is the fused feature of texture and structure. 

\subsection{Descriptor activation mapping (DAM)}   
We propose a descriptor activation mapping (DAM) to visually interpret how the neighbour points in contributing the above final descriptor extraction. The main idea of the proposed DAM is to utilize the descriptor's channel value as the loss to backpropagate into the target layer. The motivation is to use the gradient to investigate the significance of input points in generating the descriptor's channel value.

\begin{lemma}
	The feature map gradient is linearly related to the kernel gradient.
	\label{lemma1}
\end{lemma}
The lemma proof is attached in the supplement material.

\begin{figure}[htp]
	\centering
	\includegraphics[width=0.48\textwidth]{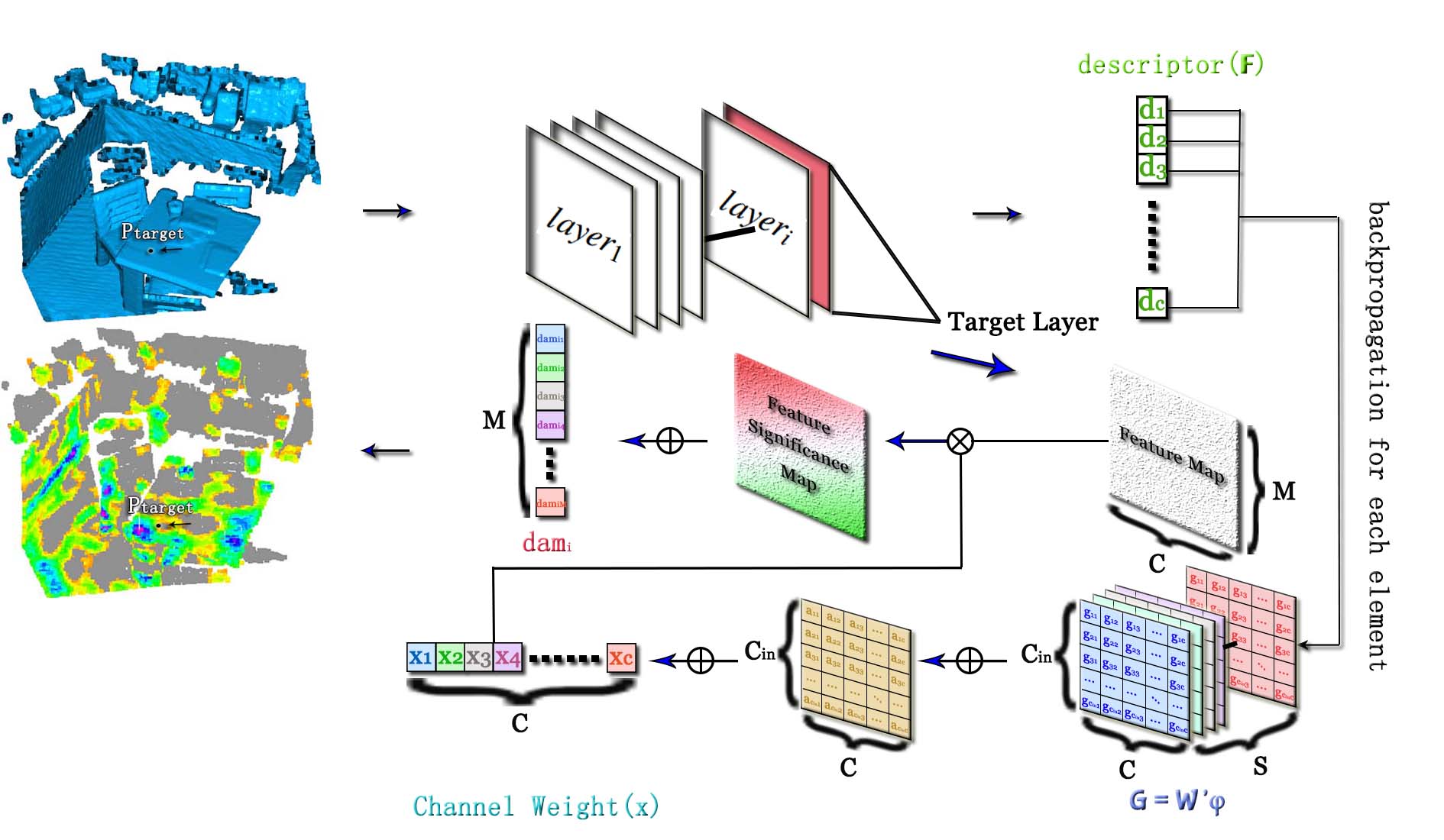}
	\caption{Diagram of the DAM. }
	\label{fig:3}
\end{figure}
In the Grad-CAM\cite{selvaraju2017grad}, the class activation mapping is calculated by the addition of feature map gradient at channel dimension. Our method is based on the sparse tensor framework, where only kernel gradient is available. It is not easy to directly calculate the feature map gradient.
According to Lemma \ref{lemma1}, the feature map gradient has a linear relation with the kernel gradient. In the proposed DAM, we introduce a method to calculate the activation mapping for the 3D point descriptor with only kernel gradient available. 

Figure \ref{fig:3} shows the calculation process. Firstly, the point descriptor $F \in \mathbb{R}^{M\times C}$ is calculated by running a forward step of the descriptor network. Secondly, we consider the descriptor's each dimension as a loss, and the loss is backpropagated from the descriptor to the last layer (target layer). After the backpropagation, we can obtain kernel gradient of $i^{th}$ descriptor element $G \in R^{ S \times C_{in} \times C}$ in the sparse tensor framework as
\begin{eqnarray}
	\begin{aligned}
		G&=\frac{\partial d_i}{\partial \omega}\varphi
	\end{aligned}
\end{eqnarray}
where $\frac{\partial d_i}{\partial \omega}$ is the gradient of $i^{th}$ descriptor element $d_i$ related to kernel parameter $\omega$, which can be obtained from sparse tensors' automatic back-propagation operation. $S$ represents the size of the convolution kernel, $C_{in}$ represents the size of the input channel, $C$ represents the size of the output channel, $\varphi$ represents a marker function, 1 if $d_i>0$, and -1 otherwise. 

Thirdly,  the channel weight of $i^{th}$ descriptor element $x \in R^{1\times C}$ is calculated by adding up the kernel gradient $G$ along the convolution kernel dimension $S$ and input channel dimension $C_{in}$.
\begin{eqnarray}
	x = \sum_{j=1}^{C_{in}}\sum_{i=1}^{S} G_{ijk}, \forall k\in[1,C],
\end{eqnarray}
The channel weight $x$ describes the significance of each channel on the output feature map of the target layer.

\begin{figure*}[h]
	\centering
	\begin{subfigure}{0.42\textwidth}
		\centering
		\includegraphics[width=0.9\linewidth, height=4cm]{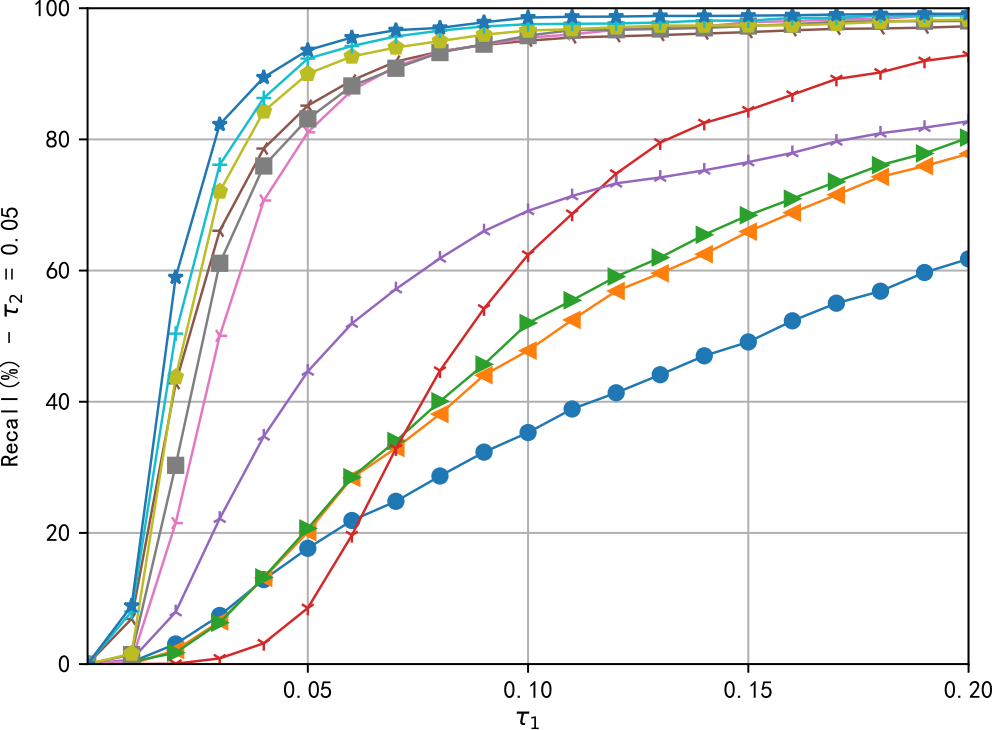} 
		\caption{Inlier Distance Threshold}
		\label{fig:subim1}
	\end{subfigure}
	\begin{subfigure}{0.42\textwidth}
		\centering
		\includegraphics[width=0.9\linewidth, height=4cm]{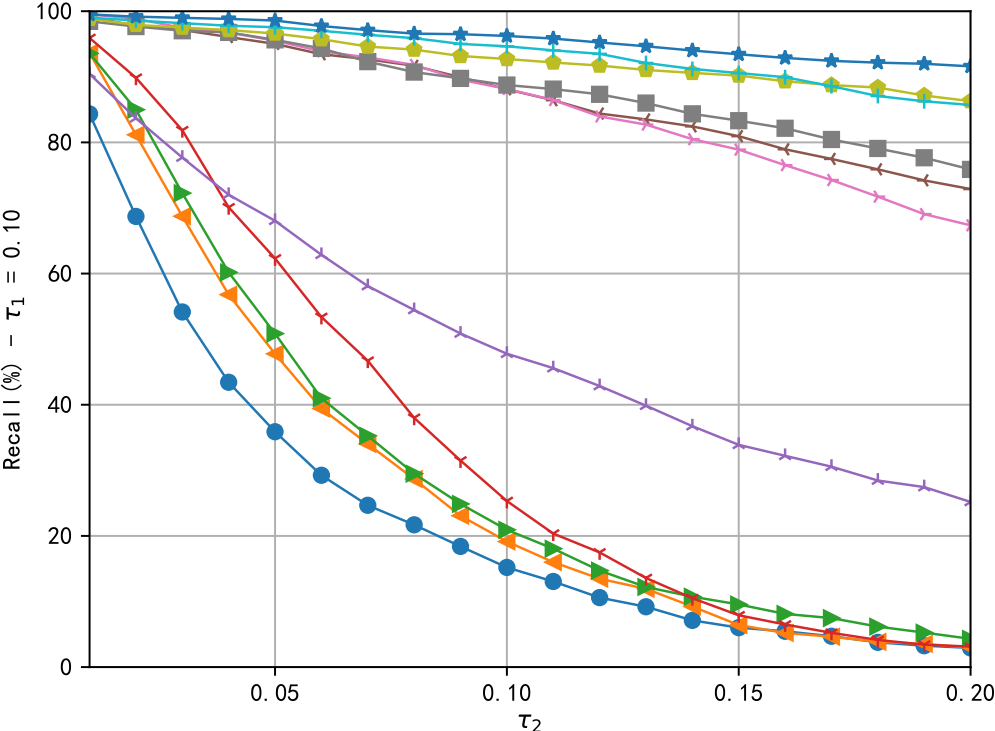}
		\caption{Inlier Ratio Threshold}
		\label{fig:subim2}
	\end{subfigure}
	\begin{subfigure}{0.15\textwidth}
		\centering
		\includegraphics[width=0.95\linewidth, height=4.4cm]{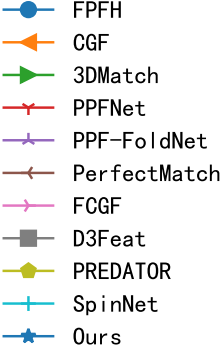}
		\label{b}
	\end{subfigure}
	\caption{Comparison of IMFNet with other state-of-the-art method on 3DMatch under different inlier distance threshold $\tau_1$(a) and different inlier ratio threshold $\tau_2$(b).}
	\label{fig:4}
\end{figure*}

Fourthly, the feature significance map of the target layer is obtained by multiplying the weights with the feature map. The feature significance map represents the significance of each feature channel on the output feature map of the target layer contributing to the final descriptor value. Then, the descriptor activation mapping from  $i^{th}$ descriptor element  is calculated by summing up the feature significance map along channel dimension. Mathematically,
\begin{eqnarray}
	dam_i  =\frac{1}{C}\sum_{i=1}^{C}(F*x)_i
\end{eqnarray}
The $dam_i \in R^{M \times 1}$ describes the contribution of input points to the value of $i^{th}$ descriptor element.

Finally, the descriptor activation map is calculated by summing $dam_i$ from all the $C$ output descriptor elements.
\begin{eqnarray}
	DAM \in R^{M \times 1} = Relu(\sum_{i=1}^{C}dam_i)
\end{eqnarray}


\section{Experiments}

The proposed algorithm is trained using the same loss and parameters of FCGF \cite{choy2019fully}. Then, following FCGF \cite{choy2019fully} and SpinNet \cite{ao2021spinnet}, we evaluate our IMFNet on the indoor 3DMatch \cite{zeng20173dmatch} and outdoor KITTI \cite{geiger2012we} datasets. We also evaluate on 3DLoMatch \cite{huang2021predator} that contains low overlap pairs between $10\%-30\%$. Regarding the 3DMatch and 3DLoMatch, we manually inspect and select the images for each point cloud to construct a dataset of paired images and point clouds named \textit{3DImageMatch}. Our experiments are conducted on this dataset. The dataset construction and training details are attached in the supplement material.

\textbf{Ground truth.}
Given pair of fragments \textbf{P} and \textbf{Q}, following FCGF\cite{choy2019fully} and SpinNet \cite{ao2021spinnet}, we randomly select 5000 anchor points from the overlapping region of \textbf{P}, and then apply the ground truth transformation \textbf{T = $\{$R,t$\}$} to determine the corresponding point in \textbf{Q} fragment. The descriptor is evaluated using these ground-truth correspondences.

\subsection{Evaluation on 3DMatch}
The 3DMatch \cite{zeng20173dmatch}  is a well-known indoor registration dataset of 62 scenes captured by RGBD sensor. Following the experimental setting of FCGF and SpinNet, we train the network and evaluate the descriptor's performance. 

\begin{figure*}[h]
	\includegraphics[width=\linewidth]{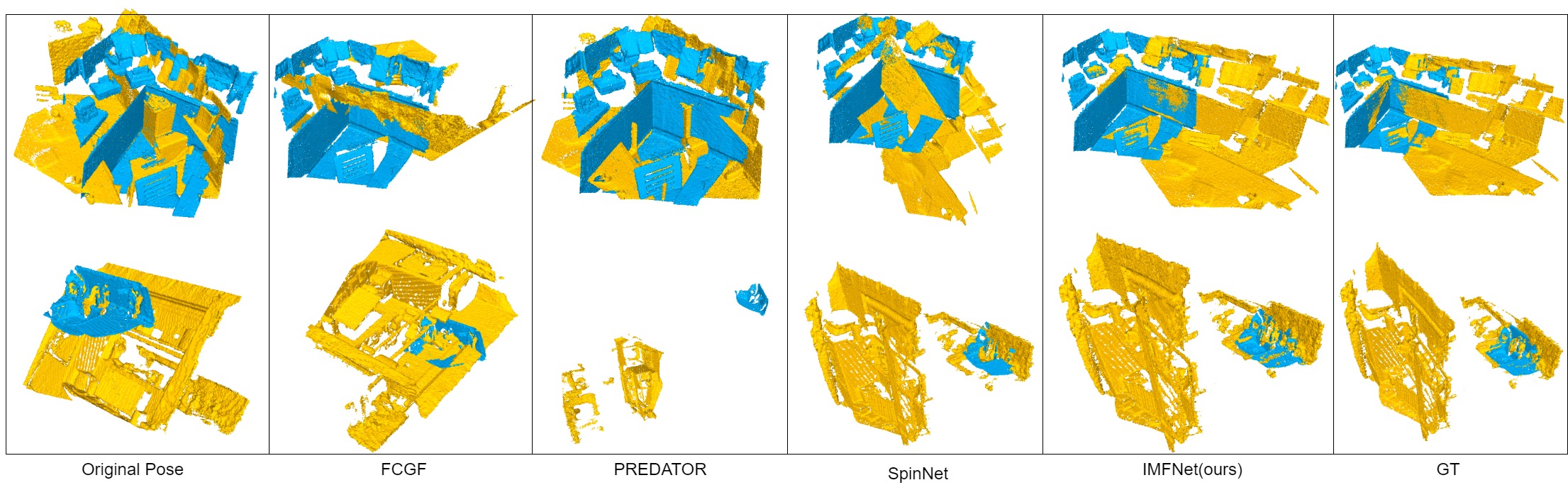}
	\caption{Visual comparison on 3DMatch dataset.}
	\label{f3dmatch}
\end{figure*}

\begin{table}[h]
	\begin{center}
		\scriptsize
		\begin{tabular}{p{2.1cm}|p{0.5cm}p{0.5cm}|p{0.5cm}p{0.5cm}p{0.5cm}p{0.5cm}}	\hline
			\multirow{2}{*}{Method}
			&\multicolumn{2}{c|}{{Origin(\%)}}
			&\multicolumn{4}{c}{{Rotated(\%)}}\\
			&$\tau_2(0.05)$                &\makecell[c]{Std} 
			&$\tau_2(0.05)$                &\makecell[r]{Std} 
			&$\tau_2(0.2)$                &\makecell[c]{Std}\\
			\hline
			FPFH\cite{rusu2009fast} 	            
			&\makecell[c]{35.9}           &\makecell[c]{13.4}     
			&\makecell[c]{36.4}           &\makecell[c]{13.6}
			&\makecell[c]{-}              &\makecell[c]{-}  \\
			CGF \cite{khoury2017learning}	                                
			&\makecell[c]{58.2}           &\makecell[c]{14.2}
			&\makecell[c]{47.8}           &\makecell[c]{14.0}
			&\makecell[c]{3.0}            &\makecell[c]{-}\\
			3DMatch\cite{zeng20173dmatch} 	        
			&\makecell[c]{59.6}           &\makecell[c]{8.8}
			&\makecell[c]{50.8}           &\makecell[c]{-}
			&\makecell[c]{4.3}            &\makecell[c]{-}\\
			PPFNet  \cite{deng2018ppf}                                                        
			&\makecell[c]{62.3}           &\makecell[c]{10.8}
			&\makecell[c]{3.1}            &\makecell[c]{-}
			&\makecell[c]{0.3}           &\makecell[c]{-}\\
			PPF-FoldNet  \cite{deng2018ppf}                           
			&\makecell[c]{71.8}           &\makecell[c]{10.5}        
			&\makecell[c]{73.1}           &\makecell[c]{10.4}
			&\makecell[c]{25.1}           &\makecell[c]{-} \\
			PerfectMatch \cite{gojcic2019perfect}                           
			&\makecell[c]{94.7}           &\makecell[c]{2.7}
			&\makecell[c]{94.9}           &\makecell[c]{2.5}
			&\makecell[c]{72.9}           &\makecell[c]{-}\\
			FCGF\cite{choy2019fully}                
			&\makecell[c]{95.2}           &\makecell[c]{2.9}
			&\makecell[c]{95.3}           &\makecell[c]{3.3}
			&\makecell[c]{67.4}           &\makecell[c]{-}\\
			D3Feat\cite{bai2020d3feat}              
			&\makecell[c]{95.8}           &\makecell[c]{2.9}
			&\makecell[c]{95.5}           &\makecell[c]{3.5}
			&\makecell[c]{75.8}           &\makecell[c]{-}\\
			LMVD \cite{li2020end}               
			&\makecell[c]{97.5}           &\makecell[c]{2.8}
			&\makecell[c]{96.9}           &\makecell[c]{-}
			&\makecell[c]{86.9}           &\makecell[c]{6.6}\\
			PREDATOR\cite{huang2021predator}        
			&\makecell[c]{-}              &\makecell[c]{-}
			&\makecell[c]{96.7}           &\makecell[c]{-}
			&\makecell[c]{86.2}           &\makecell[c]{-}\\
			SpinNet\cite{ao2021spinnet}             
			&\makecell[c]{97.6}           &\makecell[c]{1.9}
			&\makecell[c]{97.5}           &\makecell[c]{1.5}
			&\makecell[c]{85.7}           &\makecell[c]{-}\\
			MS-SVConv \cite{horache20213d}           
			&\makecell[c]{-}              &\makecell[c]{-}
			&\makecell[c]{98.4}           &\makecell[c]{-}
			&\makecell[c]{89.9}           &\makecell[c]{-}\\ 
			\hline
			\textbf{Ours}
			&\makecell[c]{\textbf{98.5}}     &\makecell[c]{\textbf{1.8}}
			&\makecell[c]{\textbf{98.6}}  &\makecell[c]{\textbf{1.5}}
			&\makecell[c]{\textbf{91.6}}  &\makecell[c]{\textbf{4.4}}\\
			\hline
			
		\end{tabular}
	\end{center}
	\caption{Feature match recall (FMR) on 3DMatch.}
	\label{t1}
\end{table}
Table \ref{t1} shows the feature match recall (FMR) comparison with the current state-of-the-art methods. The results show that the proposed IMFNet obtains state-of-the-art accuracy. Notably, our IMFNet achieves 91.6\% ($\uparrow$1.7\%) in inlier threshold $\tau_2=0.2$, which shows that multimodal fusion can reduce the outliers and improve the descriptors' distinctiveness. Figure \ref{f3dmatch} visually demonstrates the better performance than the recent state-of-the-art descriptors.

\textbf{Evaluation on different error thresholds.} 
Following the experimental setting of FCGF\cite{choy2019fully}, Figure \ref{fig:4} shows the accuracy comparison on different error thresholds ($\tau_1$ and $\tau_2$). As shown in Figure \ref{fig:4}, the proposed IMFNet consistently outperforms all other methods across different error thresholds. This experiment demonstrates that multimodal fusion improves the descriptor's distinctiveness and achieves state-of-the-art accuracy among different accuracy requirements. Looking at Figure \ref{fig:4}-(b), it is worth noting that FMR scores of our IMFNet are significantly higher than those of other methods when a high matching rate (e.g., $\tau_2 = 20\%$) is required.

\begin{table}[h]
	\begin{center}
		\scriptsize
		\begin{tabular}{p{1.8cm}|p{0.5cm}p{0.5cm}p{0.5cm}p{0.5cm}p{0.5cm}|p{0.5cm}}	\hline	
			\multirow{2}{*}{Method}       
			&\makecell[c]{\multirow{2}{*}{{5000}}}
			&\makecell[c]{\multirow{2}{*}{{2500}}}
			&\makecell[c]{\multirow{2}{*}{{1000}}}
			&\makecell[c]{\multirow{2}{*}{{500}}}
			&\makecell[c]{\multirow{2}{*}{{250}}}
			&\makecell[c]{\multirow{2}{*}{{Avg}}} \\
			& & & & & &\\
			\hline
			PerfectMatch \cite{gojcic2019perfect}
			&\makecell[c]{94.7}     &\makecell[c]{94.2}     &\makecell[c]{92.6}      
			&\makecell[c]{90.1}     &\makecell[c]{82.9}     &\makecell[c]{90.9}\\
			FCGF\cite{choy2019fully}         
			&\makecell[c]{95.2}     &\makecell[c]{95.5}     &\makecell[c]{94.6}
			&\makecell[c]{93.0}     &\makecell[c]{89.9}     &\makecell[c]{93.6}\\
			D3Feat-rand\cite{bai2020d3feat}  
			&\makecell[c]{95.3}     &\makecell[c]{95.1}     &\makecell[c]{94.2}
			&\makecell[c]{93.6}     &\makecell[c]{90.8}     &\makecell[c]{93.8}\\
			D3Feat-pred\cite{bai2020d3feat}  
			&\makecell[c]{95.8}     &\makecell[c]{95.6}     &\makecell[c]{94.6}
			&\makecell[c]{94.3}     &\makecell[c]{93.3}     &\makecell[c]{94.7}\\
			SpinNet\cite{ao2021spinnet}      
			&\makecell[c]{97.6}     &\makecell[c]{97.5}     &\makecell[c]{97.3}
			&\makecell[c]{96.3}     &\makecell[c]{94.3}     &\makecell[c]{96.6}\\ 
			MS-SVConv \cite{horache20213d}   
			&\makecell[c]{98.4}     &\makecell[c]{96.4}     &\makecell[c]{95.4}
			&\makecell[c]{95.0}     &\makecell[c]{93.0}     &\makecell[c]{95.6}\\ 
			\hline
			
			\textbf{Ours} 
			&\makecell[c]{\textbf{98.6}}        &\makecell[c]{\textbf{98.5}} 
			&\makecell[c]{\textbf{98.2}}        &\makecell[c]{\textbf{98.1}}
			&\makecell[c]{\textbf{97.5}}        &\makecell[c]{\textbf{98.2}}\\
			\hline
		\end{tabular}
	\end{center}
	\caption{Feature match recall across different number of sampled points.}
	\label{t2}
\end{table}
\textbf{Evaluation on a different number of sampled points.} 
We further evaluated the performance of IMFNet with a different number of sampled points on 3DMatch. Table \ref{t2} illustrates that the proposed IMFNet achieves state-of-the-art accuracy across the different number of sampled points. Particularly, our IMFNet achieves $>97\%$ feature match recall (FMR) across the different number of sampled points and even 3.2\% accuracy improvement on 250 sampling points. This result demonstrates that the proposed method achieves high robustness to the number of sampled points.

\textbf{Computation efficiency comparison.} 
We compared the single descriptors' extraction speed with the recent FCGF \cite{choy2019fully}, PREDATOR \cite{huang2021predator}, SpinNet \cite{ao2021spinnet} and MS-SVConv \cite{horache20213d} on the 3DMatch testing dataset. Table \ref{t3} shows that our method obtains comparable efficiency to FCGF and is faster than the recent PREDATOR, SpinNet and MS-SVConv. Specifically, the proposed method obtains $>440$ times more speed improvement than SpinNet. This is because the spherical projection is very slow, while this step is indispensable for spherical convolution. Compared to the PREDATOR, the proposed IMFNet achieves $>5$ times speed improvements. The reason is that its transformer module contains both self and cross attention and requires a pair of point clouds for each point descriptor extraction. 

\begin{table}[h]
	\scriptsize
	\begin{center}
		\begin{tabular}{p{2.4cm}|p{2.3cm}p{2.3cm}}		\hline
			{Method} 
			&\makecell[c]{{All (s)}} 
			&\makecell[c]{{Time (s)}}      \\ 
			\hline
			FCGF\cite{choy2019fully}             
			&\makecell[c]{\textbf{25.06}}          &\makecell[c]{\textbf{0.06}}     \\ 
			PREDATOR\cite{huang2021predator}     
			&\makecell[c]{762.58}                 &\makecell[c]{0.47}              \\
			SpinNet\cite{ao2021spinnet}          
			&\makecell[c]{17155.5}                 &\makecell[c]{39.62}             \\
			MS-SVConv \cite{horache20213d}         
			&\makecell[c]{41.23}                &\makecell[c]{0.10}     \\ 
			\hline
			
			\textbf{Ours}                        
			&\makecell[c]{39.98}                   &\makecell[c]{0.09}\\
			\hline
		\end{tabular}
	\end{center}
	\caption{Running speed comparison on 3DMatch.}
	\label{t3}
\end{table}

\subsection{Evaluation on 3DLoMatch.} 
We also compare the performance on 3DLoMatch \cite{huang2021predator} with the recent descriptors FCGF \cite{choy2019fully}, PREDATOR \cite{huang2021predator}, SpinNet \cite{ao2021spinnet} and MS-SVConv \cite{horache20213d}. As shown in Table \ref{t4}, the proposed IMFNet also achieves the state-of-the-art feature match recall at the registration dataset with low-overlap point clouds.  Particularly, our method achieves 80.6\% ($\uparrow$ 2\%) at the $\tau_2=0.05$ and 49.8\% ($\uparrow $4\%) accuracy at the $\tau_2=0.2$. Although we achieve best accuracy, this experiment also shows that the low-overlap problem is still a challenge for correspondence-based registration methods.

\begin{table}[h]
	\scriptsize
	\begin{center}
		\begin{tabular}{p{2.4cm}|p{1.6cm}|p{1.6cm}|p{1.2cm}}	\hline
			\multirow{2}{*}{Method}
			&\makecell[c]{FMR(\%)}
			&\makecell[c]{FMR(\%)}
			&\makecell[c]{\multirow{2}{*}{Feat dim}}\\
			&\makecell[c]{$\tau_2(0.05)$} 
			&\makecell[c]{$\tau_2(0.20)$}\\
			\hline
			FCGF\cite{choy2019fully}             
			&\makecell[c]{54.7}     &\makecell[c]{9.3}      &\makecell[c]{32}\\
			PREDATOR\cite{huang2021predator}     
			&\makecell[c]{78.6}     &\makecell[c]{-}         &\makecell[c]{32}\\
			SpinNet\cite{ao2021spinnet}          
			&\makecell[c]{74.8}     &\makecell[c]{45.8}      &\makecell[c]{32}\\ 
			MS-SVConv \cite{horache20213d}         
			&\makecell[c]{67.9}     &\makecell[c]{27.4}      &\makecell[c]{32}\\ 
			
			\hline
			\textbf{Ours}                        
			&\makecell[c]{\textbf{80.6}}        &\makecell[c]{\textbf{49.8}}  
			&\makecell[c]{\textbf{32}}\\
			\hline
		\end{tabular}
	\end{center}
	\caption{Feature match recall (FMR) on 3DLoMatch.}
	\label{t4}
\end{table}

\begin{figure*}[ht]
	\includegraphics[width=\linewidth]{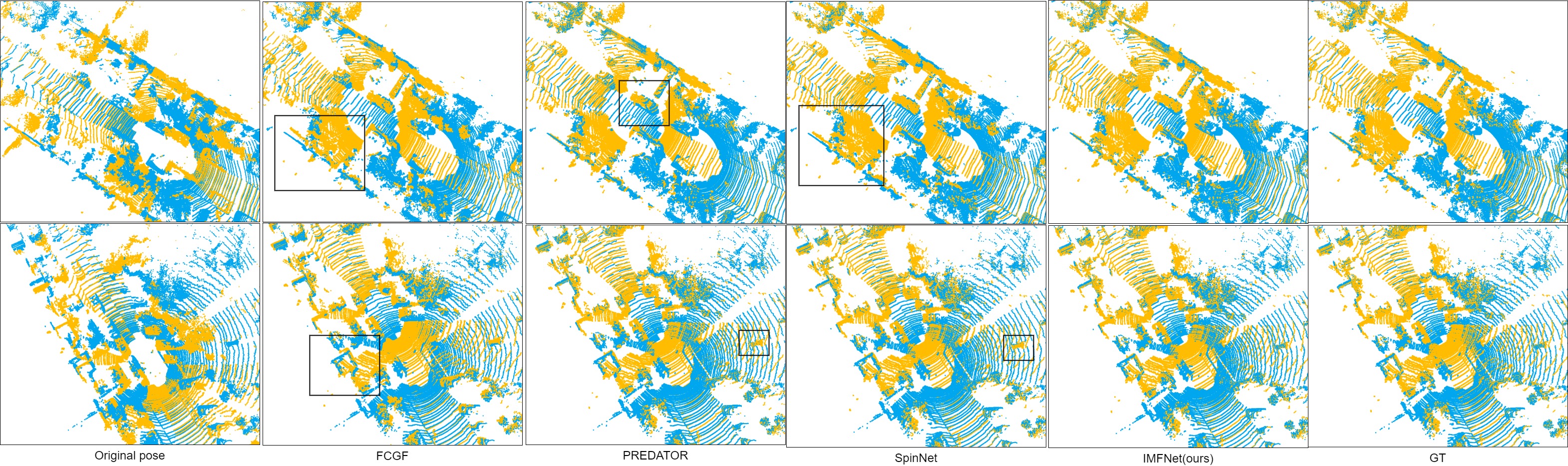}
	\caption{Visual comparison on KITTI dataset.}
	\label{fkitti}
\end{figure*}
\subsection{Evaluation on KITTI} 
KITTI \cite{geiger2012we} is well-known outdoor dataset captured by 3D LiDAR sensor. The first 11 sequences (0-10) of KITTI's odometry dataset is always used for point cloud registration evaluation because they have point cloud sequences and pose information. Following FCGF\cite{choy2019fully}, we use the first 6 sequences for training(0-5), 2 sequences for verification(6,7), and 3 sequences for testing(8-10). Following FCGF \cite{choy2019fully}, the same evaluation metrics are utilized to evaluate the relative translational error (RTE), relative rotation error (RRE), and success rate. We compare the performance of our IMFNet with the recent published descriptors FCGF \cite{choy2019fully}, D3Feat \cite{bai2020d3feat}, PREDATOR \cite{huang2021predator} and SpinNet \cite{ao2021spinnet}.

\begin{table}[h]
	\begin{center}
		\scriptsize
		\begin{tabular}{p{1.8cm}|p{0.6cm}p{0.6cm}p{0.6cm}p{0.6cm}|p{1.0cm}}		\hline
			\multirow{2}{*}{Method}  
			&\multicolumn{2}{c}{{RTE(cm)}} 
			&\multicolumn{2}{c|}{{RRE(°)}}
			&\multirow{2}{*}{{Success(\%)}}\\
			\cline{2-5}
			&\makecell[c]{Avg}  &\makecell[c]{std}  &\makecell[c]{Avg} &\makecell[c]{std}\\  
			\hline
			FCGF\cite{choy2019fully}         
			&\makecell[c]{6.47} &\makecell[c]{6.07} &\makecell[c]{0.23}&\makecell[c]{0.23} &\makecell[c]{98.92}\\
			D3Feat\cite{bai2020d3feat}       
			&\makecell[c]{6.90} &\makecell[c]{0.30} &\makecell[c]{\textbf{0.24}}    &\makecell[c]{0.06}     &\makecell[c]{\textbf{99.81}}\\
			PREDATOR\cite{huang2021predator}     
			&\makecell[c]{6.80} &\makecell[c]{-}    &\makecell[c]{0.27}&\makecell[c]{-}        &\makecell[c]{99.80}\\
			SpinNet\cite{ao2021spinnet}      
			&\makecell[c]{9.88} &\makecell[c]{0.50} &\makecell[c]{0.47}&\makecell[c]{0.09}     &\makecell[c]{99.10}\\ 
			\hline
			\textbf{Ours} 
			&\makecell[c]{\textbf{5.77}} &\makecell[c]{\textbf{0.27}} 
			&\makecell[c]{0.37} &\makecell[c]{\textbf{0.01}}
			&\makecell[c]{99.28}\\
			\hline
		\end{tabular}
	\end{center}
	\caption{Quantitative comparison on KITTI.}
	\label{t5}
\end{table}

The KITTI dataset provides both point clouds and images of the scene. However, the point clouds are 360$^\circ$ while the images only have the front view of their corresponding point clouds. The proposed IMFNet is trained and tested using the point cloud and its corresponding image. We conducted the experiments on KITTI data to demonstrate that the proposed IMFNet can improve the distinctiveness when only partial texture information is available.  The lower RTE and RRE in Table \ref{t5}  shows that the proposed IMFNet can improve the distinctiveness of point descriptors when partial texture information is available. Also, the better success rate than FCGF indicates that adding partial texture information can improve the registration performance. Figure \ref{fkitti} visually compare the registration results on the KITTI dataset.

\indent \textbf{From 3DMatch to KITTI} 
We also performed a cross-dataset evaluation on the KITTI dataset to test the generalization capability of the proposed IMFNet.  We trained all the methods on 3DMatch and tested them on KITTI with the same experimental setting (e.g., RANSAC iterates 50K times). As shown in Table \ref{t6}, our IMFNet obtains better performance than the compared methods. We also increased the RANSAC iteration time from 50K to 400K, and the proposed algorithm achieves 99.46\% registration recall. This experiment shows that multimodal fusion is a promising way to achieve a high generalization ability in cross domains.

\begin{table}[h]
	\begin{center}
		\scriptsize
		\begin{tabular}{p{1.6cm}|p{0.8cm}p{0.8cm}p{0.8cm}p{0.8cm}|p{1.0cm}}		\hline
			\multirow{2}{*}{Method}  
			&\multicolumn{2}{c}{{RTE(cm)}} 
			&\multicolumn{2}{c|}{{RRE(°)}}
			&\multirow{2}{*}{{Success(\%)}}\\
			\cline{2-5}
			&Avg       &std        &Avg       &std         \\     
			\hline
			FCGF\cite{choy2019fully}         
			&27.1      &5.58       &1.61      &1.51     &\makecell[c]{24.19}\\
			D3Feat\cite{bai2020d3feat}       
			&31.6      &10.1       &1.44      &1.35     &\makecell[c]{36.76}\\
			SpinNet\cite{ao2021spinnet}      
			&{15.6}      &{1.89}&{0.98}      
			&0.63   &\makecell[c]{81.44}\\ 
			\hline
			\textbf{Ours (50K)} 
			&21.9 &{3.10} &{1.94} &{0.30} 
			&\makecell[c]{{85.59}}\\
			\textbf{Ours(400K)} &\textbf{10.3} &\textbf{0.68} &\textbf{0.70} &\textbf{0.05} 
			&\makecell[c]{\textbf{99.46}}\\
			\hline
		\end{tabular}
	\end{center}
	\caption{The performance from 3DMatch to KITTI.}
	\label{t6}
\end{table}

\subsection{Ablation Study}
During the descriptor extraction, the critical contribution of the proposed IMFNet is the attention-fusion module. This section reported several ablation studies we have already done on this module. The ablation study is performed on the 3DMatch dataset, and the $\tau_1=0.1 (m)$ is set for all the ablation studies. 

\textbf{With/without the attention-fusion module.} Firstly, we removed the attention-Fusion module of the proposed IMFNet and extracted 3D descriptors using only structural information. After we remove the attention-fusion module, the architecture is the FCGF \cite{choy2019fully}.  Table \ref{t7} shows the feature match recall (FMR) comparison. This ablation study demonstrates that the fusion of texture information will significantly improve the feature match accuracy with a large margin. Notably, the attention-fusion module improved 24.2\% on feature match recall when the inlier threshold was set to 0.2. This ablation study demonstrates the importance of texture information in improving the descriptors' distinctiveness.

\begin{table}[h]
	\begin{center}
		\begin{tabular}{p{2.4cm}|p{2.3cm}p{2.3cm}}	
			\hline	
			\makecell[c]{Attention Fusion} 
			&\makecell[c]{$\tau_2$(0.05)}            
			&\makecell[c]{$\tau_2$(0.2)}   \\ 
			\hline
			\makecell[c]{with (w)}
			&\makecell[c]{\textbf{98.6}}            &\makecell[c]{\textbf{91.6}}  \\ 
			\makecell[c]{without (wo)}
			&\makecell[c]{95.2}                        &\makecell[c]{67.4}               \\
			\hline
		\end{tabular}
	\end{center}
	\caption{Ablation study of w/wo attention-fusion module.}
	\label{t7}
\end{table}

\indent \textbf{Single/multiple attention-fusion modules.} In the proposed IMFNet algorithm, we only use one attention-fusion module between encoder and decoder. We also added the attention-fusion module behind each upsampling layer of the decoder as additional information (three in total). This architecture could integrate hierarchical fusion between texture and structure information. The experimental results of Table \ref{t8} show that there is no significant performance increase when we consider three attention-fusion modules. However, both the GPU memory consumption and computation time largely increase. The reason is that the features obtained by the large receptive field in the last encoder layer contain enough structural information. Texture and structure fusion on these features is the most appropriate choice.

\begin{table}[h]
	\begin{center}
		
		\begin{tabular}{p{1.5cm}|c|c|c|c}	\hline
			Position  &$\tau_2(0.05)$ &$\tau_2(0.2)$ & Memory & Time(s) \\ \hline
			Single    &\textbf{98.6}  &\textbf{91.6} &$\sim$ 4GB & 0.09 \\ \hline
			Three &98.5           &91.0          &$\sim$ 6GB & 0.17\\ \hline
		\end{tabular}
	\end{center}
	\caption{Ablation study of attention-fusion modules.}
	\label{t8}
\end{table}

\textbf{Attention-fusion module design.} Our attention-fusion contains one cross-attention (CA) layer. Following the concept of Transformer architecture \cite{vaswani2017attention}, we also considered adding a certain number of self-attention (SA) layers after the cross-attention layer. Table \ref{t9} shows that the model with one-layer self-attention achieves the best accuracy. The reason is that the attention-fusion module aims to conduct feature fusion rather than feature enhancement, and multiple self-attention layers may confuse the matching relationship between structure information and texture information.

\begin{table}[h]
	\begin{center}
		\begin{tabular}{p{2.2cm}|p{2.2cm}p{2.2cm}}	
			\hline	
			\makecell[c]{{Layers}}
			&\makecell[c]{$\tau_2$(0.05)}  
			&\makecell[c]{$\tau_2$(0.2)} \\ 
			
			\hline
			\makecell[c]{CA + 0 SA}             
			&\makecell[c]{\textbf{98.6}}   &\makecell[c]{\textbf{91.6}} \\ 
			\makecell[c]{CA + 3 SA}             
			&\makecell[c]{97.9}            &\makecell[c]{89.1}          \\
			\makecell[c]{CA + 6 SA}             
			&\makecell[c]{98.0}            &\makecell[c]{90.0}          \\
			\hline
		\end{tabular}
	\end{center}
	\caption{Ablation study of different self-attention (SA) and cross-attention (CA) layers for the attention-fusion module design.}
	\label{t9}
\end{table}

\textbf{Different setting of query(Q), key(K) and value(V).} We also evaluated the different settings of Q and K for our attention-fusion module. As shown in Table \ref{t10}, the model that takes point cloud features as query $K$ and value $V$  and image features as query $Q$ achieves relatively poor feature match recall (FMR). The reason is that the output of attention-fusion will keep enhanced point feature if we consider point cloud features as $K$ and $V$. Therefore, the attention-fusion module performs a feature enhancement instead of texture and structure fusion. Moreover, the final output features only retain structural information without texture information fusion.

\begin{table}[h]
	
	\begin{center}
		\begin{tabular}{p{1.2cm}p{1.2cm}|p{2.0cm}p{2.0cm}}	
			\hline	
			\makecell[c]{{{Q}}}      
			&\makecell[c]{{{K,V}}}        
			&\makecell[c]{$\tau_1$(0.05)}
			&\makecell[c]{$\tau_2$(0.2)} \\ 
			
			\hline
			\makecell[c]{PC}                    &\makecell[c]{Image}             
			&\makecell[c]{\textbf{98.6}}         &\makecell[c]{\textbf{91.6}}\\ 
			\makecell[c]{Image}                 &\makecell[c]{PC}             
			&\makecell[c]{97.7}                     
			&\makecell[c]{84.4}            \\
			
			\hline
		\end{tabular}
	\end{center}
	\caption{Ablation study of different QKV options.}
	\label{t10}
\end{table}

\begin{figure}[ht]
	\begin{center}
		\includegraphics[width=\linewidth, height=4.8cm]{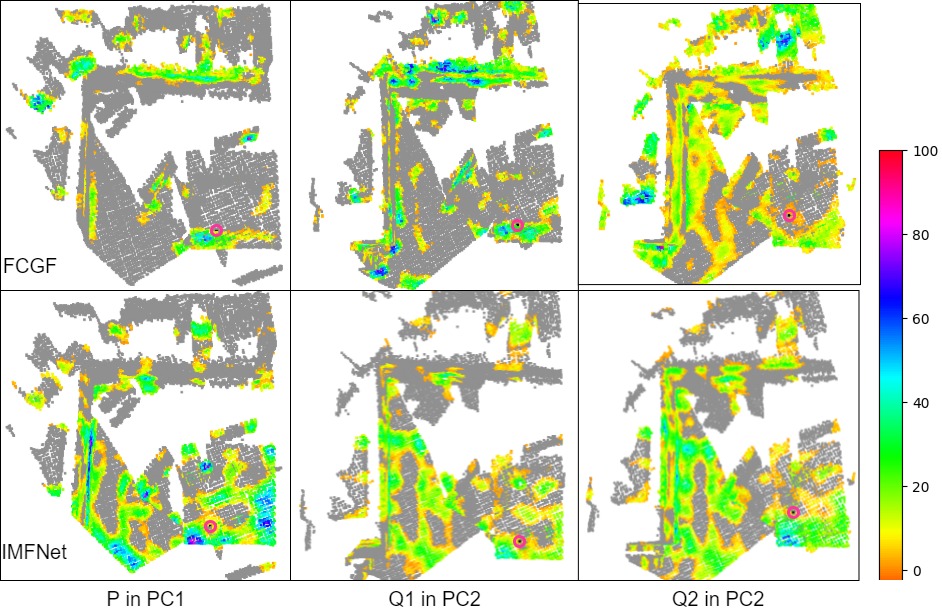}
		\caption{Interpretable results to show the point significance in generating the final descriptor. The heat maps interpret the descriptors of black points inside the red circles.}
		\label{finterpret}
	\end{center}
\end{figure}

\subsection{Interpretable results and analysis}
We conduct experiments to demonstrate the effectiveness of the proposed interpretable module. Three points $P$, $Q1$ and $Q2$ are selected based on descriptor search. $P$ and $Q1$ is matched pair, and $P$ and $Q2$ are non-matched pair. $P$ is from the one point cloud PC1, and $Q1$ and $Q2$ are from the matched point cloud PC2.  The last decoder layer is selected as the target layer. Then, we run the proposed interpretable method to interpret the proposed descriptor IMFNet, and the most related FCGF \cite{choy2019fully} on these points.

Figure \ref{finterpret} shows that descriptors of the FCGF and IMFNet have similar heat maps on the matched points ($P$ and $Q1$). In contrast, the heat maps of non-matched points ($P$ and $Q2$) show a significant difference. These results demonstrate that the proposed interpretable module can robustly generate a heat map for descriptors at different point clouds. 

Moreover, looking deep into the heat maps, our IMFNet selects relatively consistent regions to describe the selected matched points and get the similar descriptors. However, the heat maps of FCGF have variants in the matched points. Our better accuracy implies that finding the consistent important regions for the matched points is an important indicator for the descriptor's distinctiveness.  More interpretable results can be found in the supplemental materials.  

\section{Conclusion}
In this paper, we propose a simple and effective 3D descriptor by fusing structure and texture information. The experiments demonstrate that our descriptor achieves state-of-the-art accuracy and high efficiency on indoor, outdoor and low-overlap datasets. For the first time, we develop a method to move a step further in unfolding the black-box for the 3D registration tasks. The proposed interpretable module can be used to interpret neighbour points in contributing the descriptor and analyze the descriptor ability.

{\small
	\bibliographystyle{ieee_fullname}
	\bibliography{egbib}
}

\clearpage
 
\setcounter{section}{0}
In this supplementary material, we provide additional ablation study on Kitti (Sec. \ref{1}), a detailed 3DImageMatch fabrication process (Sec. \ref{2}),  data preprocessing (Sec. \ref{3}), and model training details (Sec. \ref{4}). We further provide our detail IMFNet network framework(Sec. \ref{5}) and prove the Lemma 1 (Sec. \ref{6}). Finally, we show some registration visualization(Sec. \ref{7}) and interpretable visualizations(Sec. \ref{8}).

\section{Ablation study on KITTI}
\label{1}
To evaluate the generalization ability of the attention-fusion module, we also conduct ablation studies on the KITTI dataset by using trained model of 3DMatch.

\indent \textbf{With/without the attention-fusion module.} Table \ref{tab:1} shows that  registration recall is improved  $>13\%$ if adding the attention-fusion module. This demonstrates that the texture information can largely improve the generalization ability. 
\begin{table}[h]
	\begin{center}
		\scriptsize
		\begin{tabular}{p{1.6cm}|p{0.8cm}p{0.8cm}p{0.8cm}p{0.8cm}|p{1.0cm}}	
			\hline
			\multirow{2}{*}{Attention Fusion}  &\multicolumn{2}{c}{RTE(cm)} &\multicolumn{2}{c|}{RRE(°)}
			&\multirow{2}{*}{Success(\%)}\\
			\cline{2-5}
			&Avg       &std        &Avg       &std  \\     
			\hline
			\makecell[c]{With (w)}    
			&\textbf{21.9}      &\textbf{3.10}       &\textbf{1.94}       &\textbf{0.30}     &\makecell[c]{\textbf{85.59}}\\
			\makecell[c]{Without (wo)}      
			&24.5               &3.79                &2.19                
			&0.33              &\makecell[c]{72.07}\\ 
			\hline
		\end{tabular}
	\end{center}
	\caption{Ablation study of w/wo attention-fusion module on KITTI dataset.}
	\label{tab:1}
\end{table}

\indent \textbf{Single/multiple attention-fusion modules.} In table \ref{tab:2}, the single attention-fusion module achieves the better accuracy. The reason is that the points obtained by the large receptive field in the last encoder layer contain enough structural information. Texture and structure fusion on these points is the most appropriate choice. \\
\begin{table}[h]
	\begin{center}
		\scriptsize
		\begin{tabular}{p{1.7cm}|p{0.8cm}p{0.8cm}p{0.8cm}p{0.8cm}|p{0.9cm}}	
			\hline
			\multirow{2}{*}{Attention Fusion}  &\multicolumn{2}{c}{RTE(cm)} &\multicolumn{2}{c|}{RRE(°)}
			&\multirow{2}{*}{Success(\%)}\\
			\cline{2-5}
			&Avg       &std        &Avg       &std  \\     
			\hline
			\makecell[c]{Single}     
			&\textbf{21.9}      &3.10                &\textbf{1.94}       &\textbf{0.30}     &\makecell[c]{\textbf{85.59}}\\
			\makecell[c]{Three}  
			&22.5               &\textbf{2.99}       &1.97                          &0.33              &\makecell[c]{80.36}\\ 
			\hline
			
		\end{tabular}
	\end{center}
	\caption{Ablation study of single/multiple attention-fusion modules on KITTI dataset.}
	\label{tab:2}
\end{table}


\begin{table}[h]
	\begin{center}
		\scriptsize
		\begin{tabular}{p{1.2cm}p{0.5cm}|p{0.6cm}p{0.6cm}p{0.6cm}p{0.6cm}|p{0.9cm}}	
			\hline
			\multirow{2}{*}{Image Size} 
			&\multirow{2}{*}{deep}  
			&\multicolumn{2}{c}{RTE(cm)} 
			&\multicolumn{2}{c|}{RRE(°)}
			&\multirow{2}{*}{Success(\%)}\\
			\cline{3-6}
			&        &Avg           &std          &Avg           &std\\     
			\hline
			\makecell[c]{160$\times$160}    &\makecell[c]{1}       
			&21.9          &3.10         &1.94          
			&0.30                   &\makecell[c]{\textbf{85.59}}\\
			\makecell[c]{160$\times$160}    &\makecell[c]{4}       
			&\textbf{21.7}  &3.62         &\textbf{1.73}
			&\textbf{0.23}          &\makecell[c]{84.32}\\
			\makecell[c]{160$\times$160}    &\makecell[c]{7}       
			&22.4          &\textbf{3.07}&1.87          
			&0.26                   &\makecell[c]{81.98}\\
			\hline
		\end{tabular}
	\end{center}
	\caption{Ablation study of different self-attention (SA) and cross-attention (CA) layers for the attention-fusion module design on KITTI dataset.}
	\label{tab:3}
\end{table}

\indent \textbf{Attention-fusion module design.} Table \ref{tab:3} show that the one-layer attention-fusion design achieves best success rate, which demonstrates it best generalization ability in this design. As discussed in section 4.4 of the main manuscript, increasing the number of layers of attention-fusion may confuse the matching relationship between structure information and texture information.\\

\indent \textbf{Different setting of query(Q), key(K) and value(V).} Table \ref{tab:4} shows that setting the point cloud as query achieves the best accuracy. In addition, the generalization ability of the setting point cloud as key and value matrix achieves only slightly higher than that of the model without attention-fusion module. The reason is that the attention-fusion module performs a feature enhancement when we setting as key and value, instead of texture and structure fusion. Therefore, the final output descriptor only retains structural information without texture information fusion.


\begin{table}[h]
	\begin{center}
		\scriptsize
		\begin{tabular}{p{0.8cm}p{0.8cm}|p{0.65cm}p{0.65cm}p{0.65cm}p{0.65cm}|p{0.9cm}}	
			\hline
			\multirow{2}{*}{Q} 
			&\multirow{2}{*}{K,V}  
			&\multicolumn{2}{c}{RTE(cm)} 
			&\multicolumn{2}{c|}{RRE(°)}
			&\multirow{2}{*}{Success(\%)}\\
			\cline{3-6}
			&       &Avg       &std        &Avg       &std  \\     
			\hline
			PC      &Image  &\textbf{21.9}      &\textbf{3.10}       &\textbf{1.94}                              &\textbf{0.30}      &\textbf{85.59}\\
			Image   &PC     &{24.5}             &4.04                &2.15                                       &0.31               &73.33\\
			\hline
		\end{tabular}
	\end{center}
	\caption{Ablation study of different QKV options on KITTI dataset.}
	\label{tab:4}
\end{table}


\section{3DImageMatch}
\label{2}
\begin{figure*}[htp]
	\centering
	\includegraphics[width=1.0\textwidth]{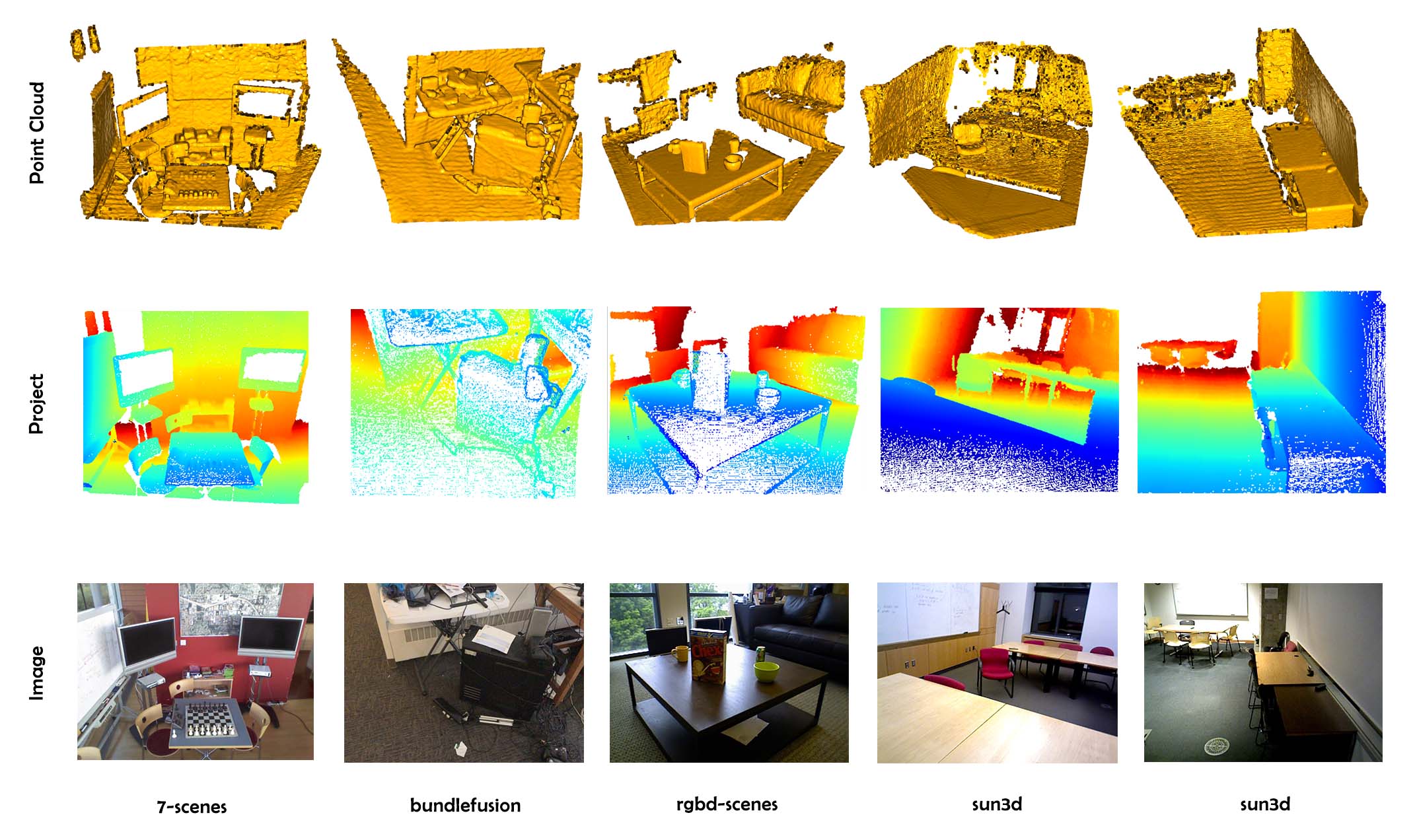}
	\caption{Some visual examples of the 3DImageMatch. The first row shows the point clouds. The second row shows projection results from point clouds to images. The third column shows the images that corresponding to the point clouds in the first row. It can be seen that the selected images are almost identical with the projected results.}
	\label{fig:1}
\end{figure*}

We consider that if we want to extract descriptors by fusing the structural information and texture information of points, we need to have a dataset of point cloud and image pairs depicting the same scene. However, there is no such dataset that contains paired point cloud and image depicting the same scene.  In this paper, we construct a dataset based on 3DMatch, named as 3DImageMatch, that consists of paired point cloud and image describing the same scene. Figure \ref{fig:1} shows several examples.

\indent \textbf{Tranining set.} In the 3DMatch training dataset, each point cloud is generated by fusing 50 depth images. To get the corresponding image of a point cloud, we need to select an image from the 50 RGB images corresponding to the 50 depth images. Since these 50 images contain slight movement, we manually select the image that has the most similar  content with the image projected by the point cloud according to the Z axis. Formally,
\[
C =\begin{bmatrix}
	f_x         &           0           &           c_x \\
	0           &           f_y         &           c_y \\
	0           &           0           &           1   \\
\end{bmatrix}
\]

\begin{eqnarray}
	\begin{aligned}
		u=f_x\frac{X}{Z}+C_x, v=f_y\frac{Y}{Z}+C_y
	\end{aligned}
\end{eqnarray}
where $C$ is the intrinsic matrix of camera, $(X, Y, Z)$ are the coordinates of a 3D point, $(u, v)$ are the coordinates of the projection point in pixels, $(cx, cy)$ is a principal point that is usually at the image center, $f_x$ and $f_y$ are the focal lengths of the camera in the $X$ and $Y$ directions, respectively. 

\indent \textbf{Test Set.} In 3DMatch's test set, we find that the point clouds  are generated from depth images as the same as those in trainning set excepting the following three scenes: 
\[7-scenes-eedkitchen\]
\[sun3d-home\_at-home\_at\_scan1\_2013\_jan\_1\]
\[Sun3d-home\_md-home\_md\_scan9\_2012\_sep\_30\]
In the above three scenarios, each point cloud is generated by fusing the first 50 depth images every 100 depth images (skip the next 50 depth images).

\section{Data preprocessing}
\label{3}
\textbf{3DMatch.} To demonstrate the value of texture information, our experiments are conducted on 3DImageMatch by following the same processing of 3DMatch. We used 54 scenarios as training sets, 6 scenarios as validation sets, and 8 scenarios as test sets. We use TSDF volumetric fusion to synthesize 50 depth images into a point cloud, and apply certain downsampling on all point clouds.\\
\indent \textbf{KITTI.} Following the setting of FCGF, we used the first 10 scenarios for model evaluation and training, among which 0-5 sequences were used for training, 6 and 7 sequences were used for verification, and 8-10 sequences were used for testing.

\section{Details of model training}
\label{4}
Since our IMFNet framework is based on FCGF, most parameters of IMFNet can refer to FCGF. We trained 200 epochs on both 3DMatch and KITTI. All of our models trained are based on $batch\_size=2$. The ResNet34 model in IMFNet is pre-trained and the image input size is $(120,160)$ on 3DMatch and $(160,160)$ on KITTI. In attention-fusioon, we set the size of the $C_t$ to be half of the dimension of the input point cloud ,i.e. $\frac{M_4}{2}$, and in the cross-attention calculation, we set only one head. We will release our code for these details.

\section{Detailed network framework}
\label{5}
\begin{figure*}[htp]
	\centering
	\includegraphics[width=1.0\textwidth]{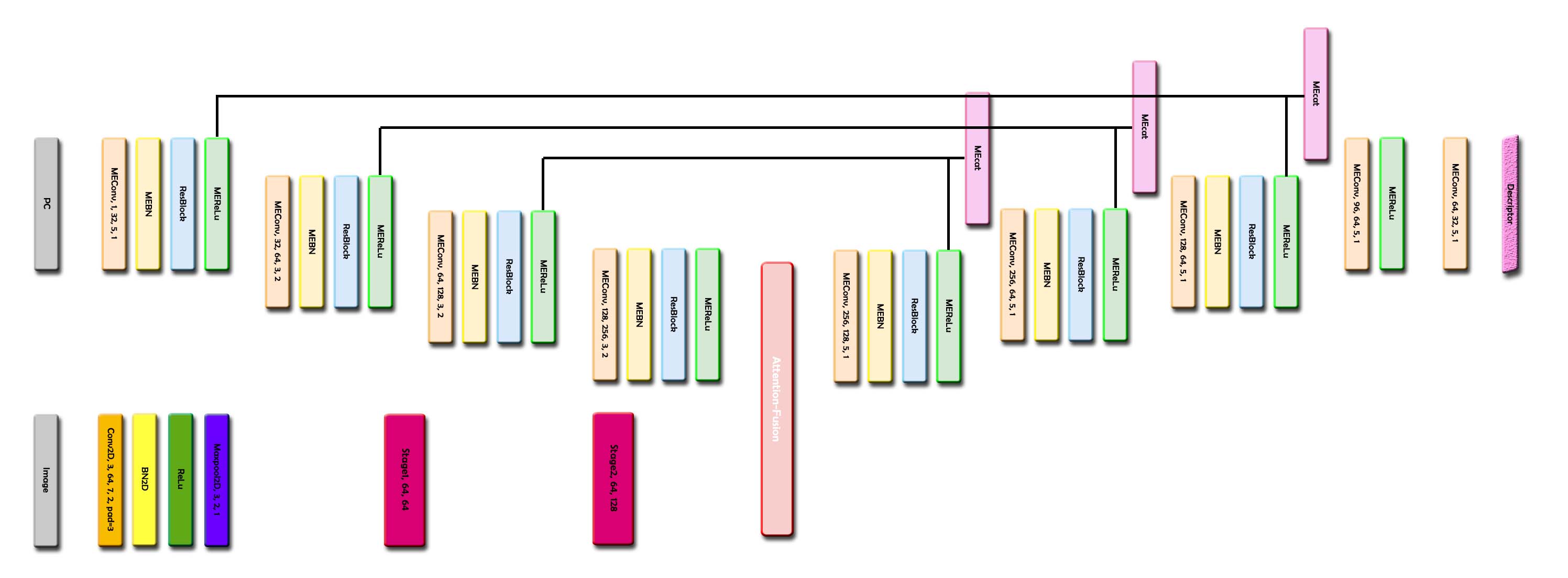}
	\caption{Detailed network framework of the proposed IMFNet.}
	\label{fig:2}
\end{figure*}

\begin{figure*}[htp]
	\centering
	\includegraphics[width=1.0\textwidth]{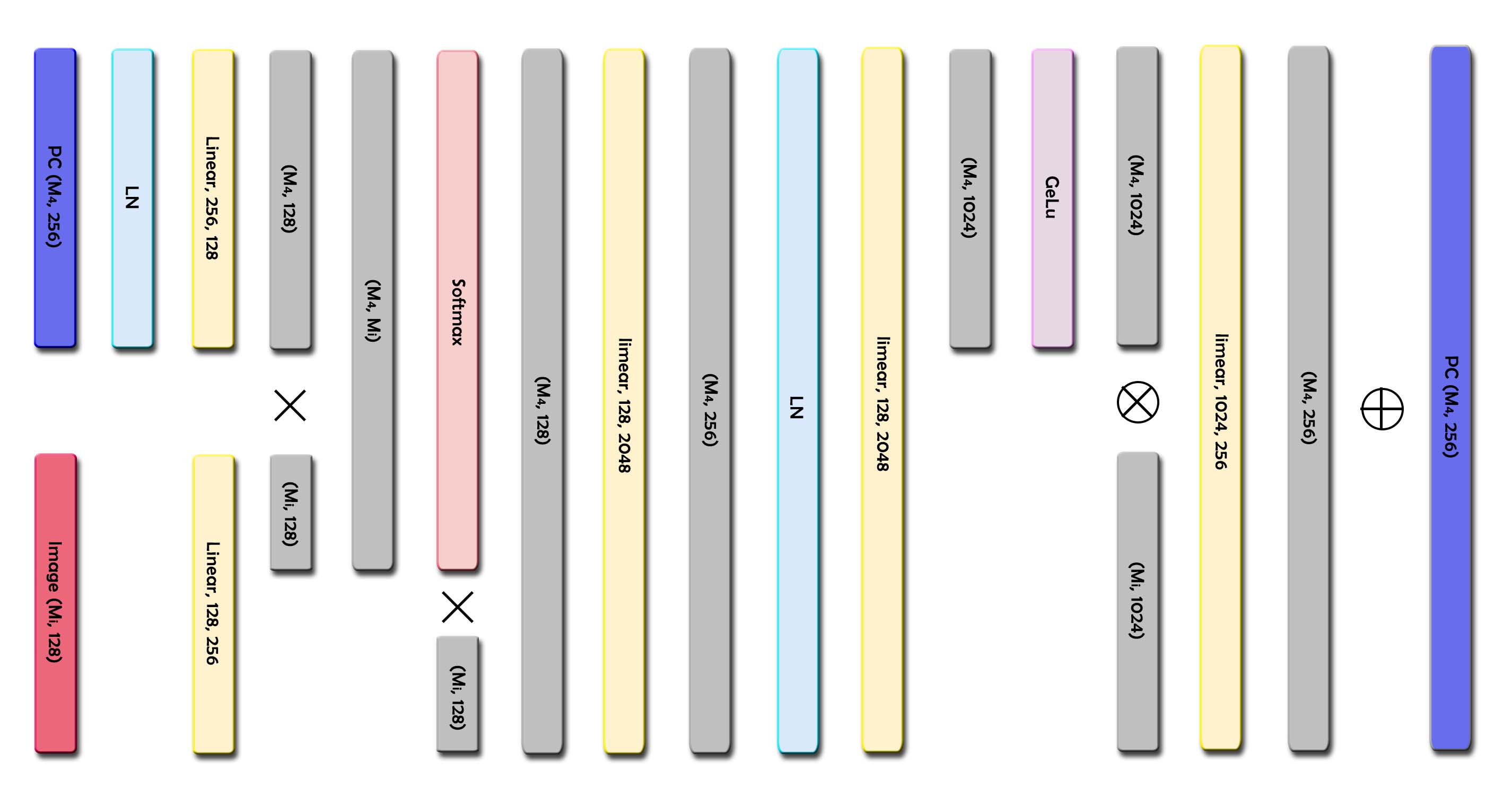}
	\caption{The details of attention-fusion module. $\times$ stands for matrix multiplication. $\otimes$ means the element multiplication, and $\oplus$ means the element addition.}
	\label{fig:3}
\end{figure*}

The details of our IMFNet network framework are shown in Figure \ref{fig:2}. In this framework, the MEConv, MEBN, MEResBlock, MEReLU and MEcat refer to FCGF. Here, MEConv, MEBN, MEResBlock, MEReLU, MEcat means MinkowskiConvolution, MinkowskiBatchNorm, BasicBlockIN, relu activation function in Minkowski engine, concatenation in Minkowski engine.  For the stage of ImageEncoder, see ResNet34 \cite{he2016deep}. The details of the attention-fusion module is shown in figure \ref{fig:3}, which shows that our attention-fusion module is simple and can be implemented easily.

\section{Proof of Lemma 1}
\label{6}
Define the input feature map of a target layer as $A \in R^{M \times C_{in}}$, the output feature map as $Z \in R^{M \times C_{out}}$, the Kernal as $K \in R^{C_{in} \times C_{out}}$ , and the kernal size as $[1,1,1]$. We perform an element-wise addition for the above $n^3$ output feature maps if kernel size is $[n,n,n]$. In forward propagation:
\begin{eqnarray}
	\begin{aligned}
		&A \times K = Z \\
		&Z_{ij}=\sum_{n=1}^{C_{in}}A_{in}K_{nj}, \forall i \in M , \forall j \in C_{out}
	\end{aligned}
\end{eqnarray}
where $C_{in}$ represents the input dimension of the target layer, $C_{out}$ represents the output dimension of the target layer, and M represents the number of elements. Let $\epsilon$ is the loss that we want to propagate backwards. Then, Kernel gradient was calculated by backward propagation as:

\begin{eqnarray}
	\begin{aligned}
		\frac{\partial \epsilon}{\partial K_{ij}}=\frac{\partial \epsilon}{\partial Z}\frac{\partial Z}{\partial K_{ij}}
	\end{aligned}
\end{eqnarray}

due to the:

\begin{eqnarray}
	\begin{aligned}
		&Z_{ij}=\sum_{n=1}^{C_{in}}A_{in}K_{nj},\forall i \in M , \forall j \in C_{out}\\
		&\frac{\partial Z_{nj}}{\partial K_{ij}} = A_{nj},\forall n \in M
	\end{aligned}
\end{eqnarray}

so:
\begin{eqnarray}
	\begin{aligned}
		&\frac{\partial \epsilon}{\partial K_{ij}}=\frac{\partial \epsilon}{\partial Z_{1j}}\frac{\partial Z_{1j}}{\partial K_{ij}}+...+\frac{\partial \epsilon}{\partial Z_{Mj}}\frac{\partial Z_{Mj}}{\partial K_{ij}}\\
		&\frac{\partial \epsilon}{\partial K_{ij}}=\sum_{n=1}^{M}\frac{\partial \epsilon}{\partial Z_{nj}}A_{nj},\forall j \in C_{out}
	\end{aligned}
	\label{kij}
\end{eqnarray}

The above equation \ref{kij} shows that the kernel gradient $\frac{\partial \epsilon}{\partial K_{ij}}$ is only related to the same $j$ column (channel dimension) of output feature map gradient $\frac{\partial \epsilon}{\partial Z_{nj}}$. Because $A_{nj}$ is scalar, equation \ref{kij} means that kernel gradient is linearly related to the output feature map gradient  in the same channel. Therefore, the Lemma 1 has been proved.

\section{Interpretable visualization}
\label{7}
\begin{figure*}[htp]
	\centering
	\includegraphics[width=1.0\textwidth, height=21cm]{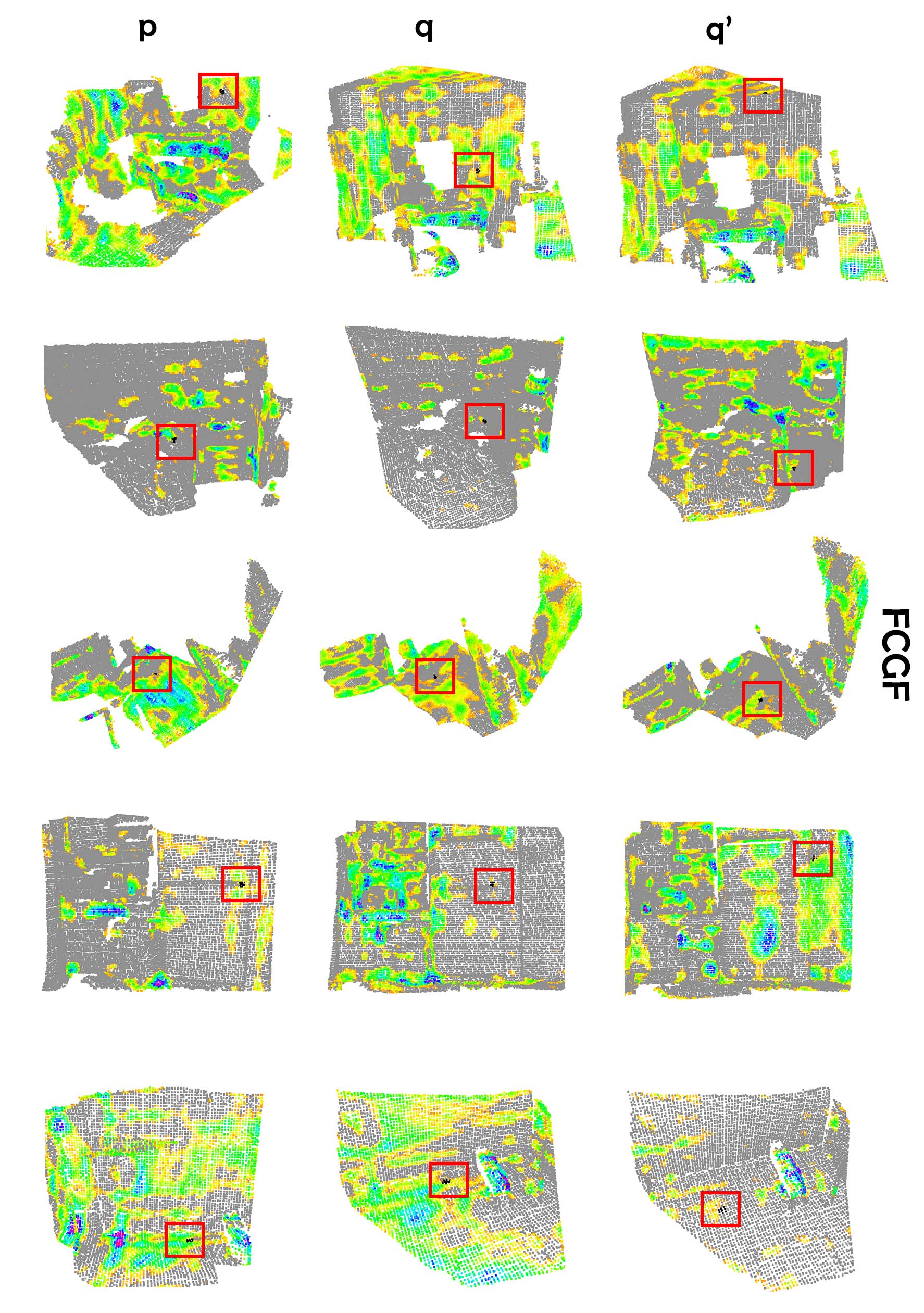}
	\caption{Interpretable results of FCGF. Points \textbf{p} and \textbf{q} are matched, and \textbf{p} and \textbf{q'} are non-matched.}
	\label{fig:4}
\end{figure*}
\begin{figure*}[htp]
	\centering
	\includegraphics[width=1.0\textwidth, height=21cm]{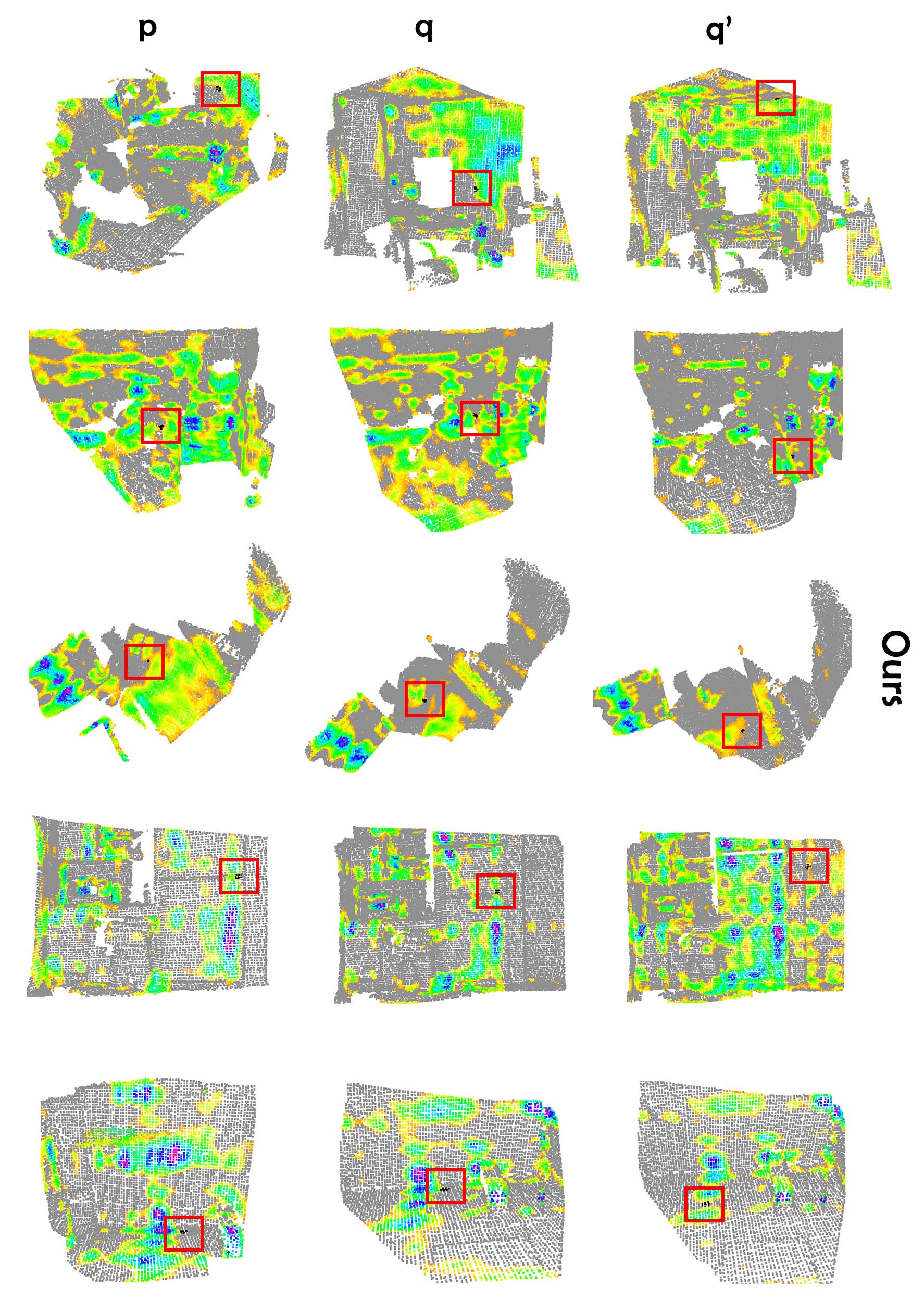}
	\caption{Interpretable results of our IMFNet. Points \textbf{p} and \textbf{q} are matched, and \textbf{p} and \textbf{q'} are non-matched.}
	\label{fig:5}
\end{figure*}
In this section, we show more interpretable results for both FCGF (figure \ref{fig:4}) and our IMFNet (figure \ref{fig:5}). Inside each heat map, the black point region is generated by using KNN to find the nearest 10 neighbor points around the target point. Both Figure \ref{fig:4} and Figure \ref{fig:5} illustrate that heat maps of matched points are similar while different in the non-matched points.  Compared with the heat maps of FCGF descriptor, our IMFNet shows more consistency.

\section{3DMatch/3DLoMatch visualization}
\label{8}

\begin{figure*}[htp]
	\centering
	\includegraphics[width=1.0\textwidth]{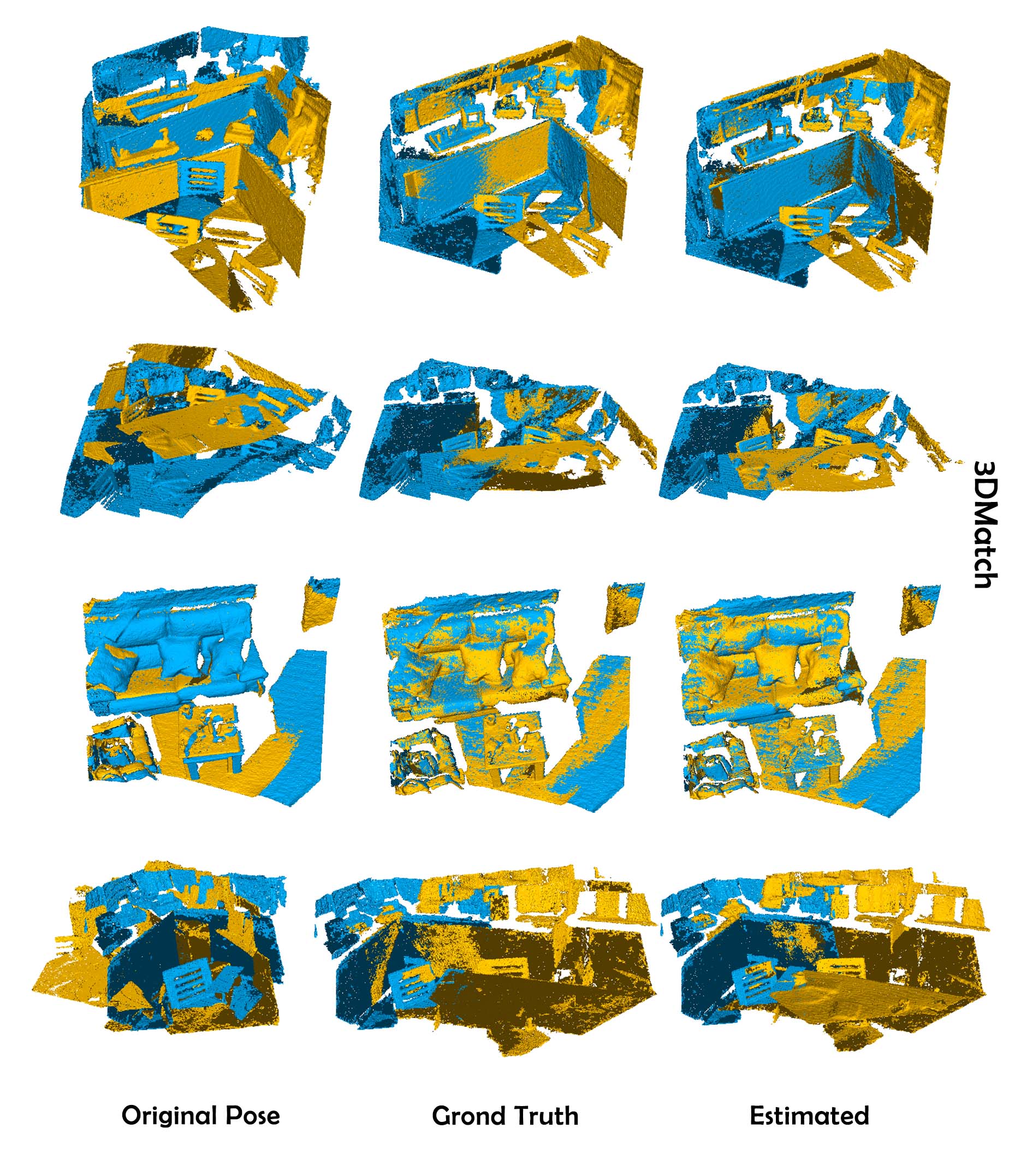}
	\caption{Some visualization results of IMFNet on 3DMatch.}
	\label{fig:6}
\end{figure*}

\begin{figure*}[htp]
	\centering
	\includegraphics[width=1.0\textwidth]{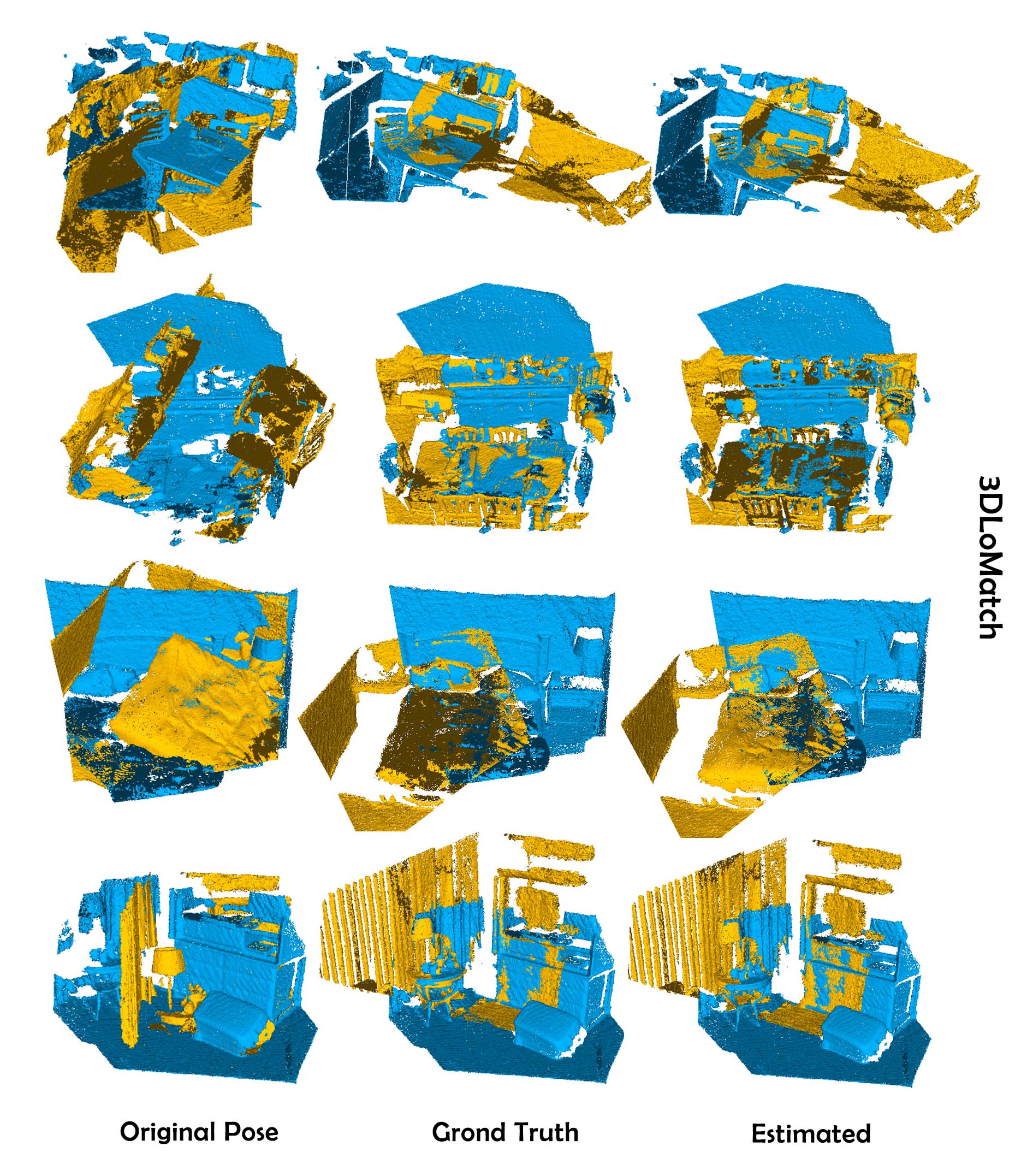}
	\caption{Some visualization results of IMFNet on 3DLoMatch.}
	\label{fig:7}
\end{figure*}
In this section, we generate more visualization examples for our IMFNet to show its registration ability. Figure \ref{fig:6} shows the visual examples on 3Dmatch, and Figure \ref{fig:7} shows the visual examples on 3DLoMatch.

\end{document}